\documentclass[10pt,twocolumn,letterpaper]{article}

\usepackage{iccv}
\usepackage{times}
\usepackage{epsfig}
\usepackage{graphicx}
\usepackage{amsmath}
\usepackage{amssymb}
\usepackage{color}
\usepackage{pgfplots}
\usepackage{verbatim}
\usepackage[percent]{overpic}

\pgfplotsset{compat=1.3}

\definecolor{dgreen}{rgb}{0,0,0}
\definecolor{dyellow}{rgb}{.7,.7,0}
\definecolor{dred}{rgb}{1,0,0}
\definecolor{dblue}{rgb}{0,0,0.7}
\definecolor{alexey}{rgb}{0.7,0,1}

\newcommand{\tcb}[1]{\textcolor{dred}{#1}}    
\newcommand{\tcn}[1]{\textcolor{dgreen}{#1}}    

\newcommand{\ognet}{OGN}

\newcommand{\shapenetcars}{ShapeNet-cars}
\newcommand{\shapenetall}{ShapeNet-all}
\newcommand{\blendswap}{BlendSwap}

\usepackage[pagebackref=true,breaklinks=true,letterpaper=true,colorlinks,bookmarks=false]{hyperref}

\iccvfinalcopy 

\def\iccvPaperID{828} 

\ificcvfinal\fi
\begin{document}

\title{Octree Generating Networks:\\ Efficient Convolutional Architectures for High-resolution 3D Outputs}

\author{Maxim Tatarchenko\textsuperscript{1} \qquad\qquad  Alexey Dosovitskiy\textsuperscript{1,2}  \qquad\qquad Thomas Brox\textsuperscript{1}\\
	{\tt\small tatarchm@cs.uni-freiburg.de \qquad adosovitskiy@gmail.com \qquad brox@cs.uni-freiburg.de}\\
	\textsuperscript{1}University of Freiburg \qquad \textsuperscript{2}Intel Labs}

\maketitle
\thispagestyle{empty}

\begin{abstract}
We present a deep convolutional decoder architecture that can generate volumetric 3D outputs in a compute- and memory-efficient manner by using an octree representation. The network learns to predict both the structure of the octree, and the occupancy values of individual cells. This makes it a particularly valuable technique for generating 3D shapes. In contrast to standard decoders acting on regular voxel grids, the architecture does not have cubic complexity. This allows representing much higher resolution outputs with a limited memory budget. We demonstrate this in several application domains, including 3D convolutional autoencoders, generation of objects and whole scenes from high-level representations, and shape from a single image. 
\end{abstract}

\section{Introduction}

Up-convolutional\footnote{Also known as deconvolutional} decoder architectures have become a standard tool for tasks requiring image generation~\cite{chairs_pami, dcgan, Johnson2016_superresolution} or per-pixel prediction~\cite{Noh2016_segm, flownet}.
They consist of a series of convolutional and up-convolutional (upsampling+convolution) layers operating on regular grids, with resolution gradually increasing towards the output of the network.
The architecture is trivially generalized to volumetric data.
However, because of cubic scaling of computational and memory requirements, training up-convolutional decoders becomes infeasible for high-resolution three-dimensional outputs.

Poor scaling can be resolved by exploiting structure in the data.
In many learning tasks, neighboring voxels on a voxel grid share the same state~--- for instance, if the voxel grid represents a binary occupancy map or a multi-class labeling of a three-dimensional object or a scene.
In this case, data can be efficiently represented with octrees~--- data structures with adaptive cell size.
Large regions of space sharing the same value can be represented with a single large cell of an octree, resulting in savings in computation and memory compared to a fine regular grid.
At the same time, fine details are not lost and can still be represented by small cells of the octree.

\begin{figure}[t]
\begin{center}
\begin{overpic}[width=\linewidth]{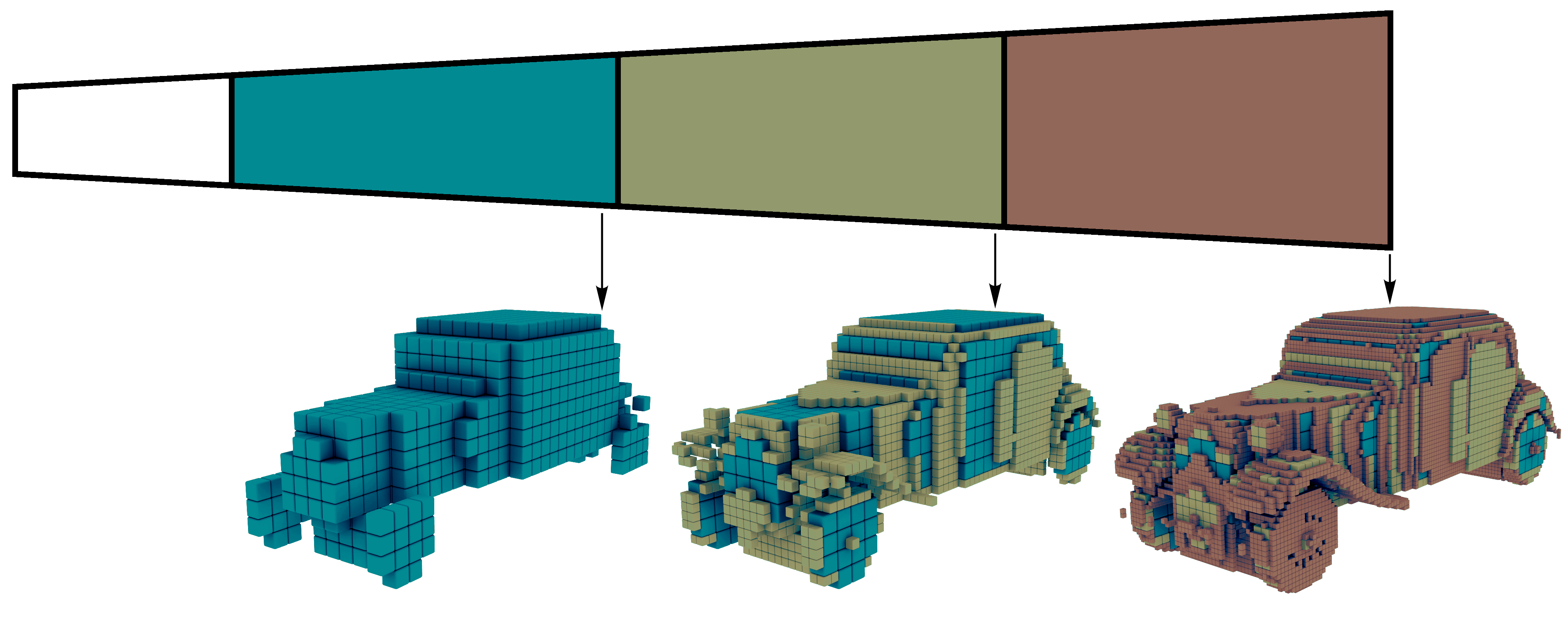}
 \put (4,30) {\footnotesize dense}
 \put (23,32) {\color{white}\footnotesize Octree}
 \put (23.2,28.3) {\color{white}\footnotesize level 1}
 \put (48,32) {\color{white}\footnotesize Octree}
 \put (48.2,28.3) {\color{white}\footnotesize level 2}
 \put (72,32) {\color{white}\footnotesize Octree}
 \put (72,28.3) {\color{white}\footnotesize level 3}
 \put (24,-2) {\footnotesize $32^3$}
 \put (52,-2) {\footnotesize $64^3$}
 \put (81,-2) {\footnotesize $128^3$}
\end{overpic}
\end{center}
\caption{The proposed \ognet{} represents its volumetric output as an octree. Initially estimated rough low-resolution structure is gradually refined to a desired high resolution. At each level only a sparse set of spatial locations is predicted. This representation is significantly more efficient than a dense voxel grid and allows generating volumes as large as $512^3$ voxels on a modern GPU in a single forward pass.}
\label{fig:teaser}
\end{figure}
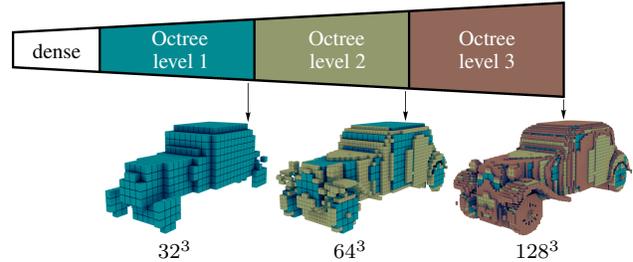

We present an octree generating network (\ognet{}) - a convolutional decoder operating on octrees.
The coarse structure of the network is illustrated in Figure~\ref{fig:teaser}.
Similar to a usual up-convolutional decoder, the representation is gradually convolved with learned filters and up-sampled.
The difference is that, starting from a certain layer in the network, dense regular grids are replaced by octrees.
Therefore, the \ognet{} predicts large uniform regions of the output space already at early decoding stages, saving the computation for the subsequent high-resolution layers.
Only regions containing fine details are processed by these more computationally demanding layers.

In this paper, we focus on generating shapes represented as binary occupancy maps.
We thoroughly compare \ognet{}s to standard dense nets on three tasks: auto-encoding shapes, generating shapes from a high-level description, and reconstructing 3D objects from single images.
{\ognet{}}s yield the same accuracy as conventional dense decoders while consuming significantly less memory and being much faster at high resolutions.
For the first time, we can generate shapes of resolution as large as $512^3$ voxels in a single forward pass.
Our OGN implementation is publicly available\footnote{\url{https://github.com/lmb-freiburg/ogn}}.

\section{Related work}

The majority of deep learning approaches generate volumetric data based on convolutional networks with feature maps and outputs represented as voxel grids. 
Applications include single- and multi-view 3D object reconstruction trained in supervised \cite{Girdhar16b, choy20163d, Di2016, GrantKG16} and unsupervised \cite{persp_trans, GadelhaMW16, NIPS2016_6600} ways,  probabilistic generative modeling of 3D shapes~\cite{Wu2015_3dshapenets, 3dgan, SharmaGF16},
semantic segmentation \cite{cicek2016_3dunet, voxresnet}
and shape deformation \cite{yumer2016learning}.
A fundamental limitation of these approaches is the low resolution of the output.
Memory and computational requirements of approaches based on the voxel grid representation scale cubically with the output size.
Thus, training networks with resolutions higher than $64^3$ comes with memory issues on the GPU or requires other measures to save memory, such as reducing the batch size or generating the volume part-by-part. 
Moreover, with growing resolution, training times become prohibitively slow.

Computational limitations of the voxel grid representation led to research on alternative representations of volumetric data in deep learning.
Tatarchenko et al. \cite{TDB16a} predict RGB images and depth maps for multiple views of an object, and fuse those into a single 3D model. This approach is not trainable end to end because of the post-processing fusion step, and is not applicable to objects with strong self-occlusion.
Sinha et al.~\cite{Sinha2016} convert shapes into two-dimensional geometry images and process those with conventional CNNs~-- an approach only applicable to certain classes of topologies.
Networks producing point clouds have been applied to object generation \cite{fan16} and semantic segmentation \cite{PointNet}.
By now, these architectures have been demonstrated to generate relatively low-resolution outputs.
Scaling these networks to higher resolution is yet to be explored.
Tulsiani et al. \cite{abstractionTulsiani17} assemble objects from volumetric primitives.
Yi et al.~\cite{SynSpecCnn} adapt the idea of graph convolutions in the spectral domain to semantic segmentation of 3D shapes.
Their approach requires all samples to have aligned eigenbasis functions, thus limiting possible application domains.

Promising alternative representations that are not directly applicable to generating 3D outputs have been explored on analysis tasks.
Masci et al.~\cite{MasBosBroVan15} proposed geodesic CNNs for extracting local features in non-Euclidean domains.
Our approach is largely inspired by Graham's sparse convolutional networks ~\cite{Graham14, Graham15}, which enable efficient shape analysis by storing 
a sparse set of non-trivial features instead of dense feature maps.
The \ognet{} essentially solves the inverse problem by predicting which regions of the output contain high-resolution information and by restricting extensive calculations only to those regions.

The recent pre-print by Riegler et al. \cite{octnet} builds on the same general idea as our work: designing convolutional networks that operate on octrees instead of voxel grids.
However, the implementation and the application range of the method is very different from our work.
When generating an octree, Riegler et al. assume the octree structure to be known at test time.
This is the case, for example, in semantic segmentation, where the structure of the output octree can be set to be identical to that of the input. 
However, in many important scenarios~--- any kind of 3D reconstruction, shape modeling, RGB-D fusion, superresolution~--- the structure of the octree is not known in advance and must be predicted. The method of Riegler et al. is not applicable in these cases.
Moreover, the \ognet{} is more flexible in that it allows for octrees with an arbitrary number of levels.

\begin{figure*}
\begin{center}
\begin{overpic}[width=\linewidth]{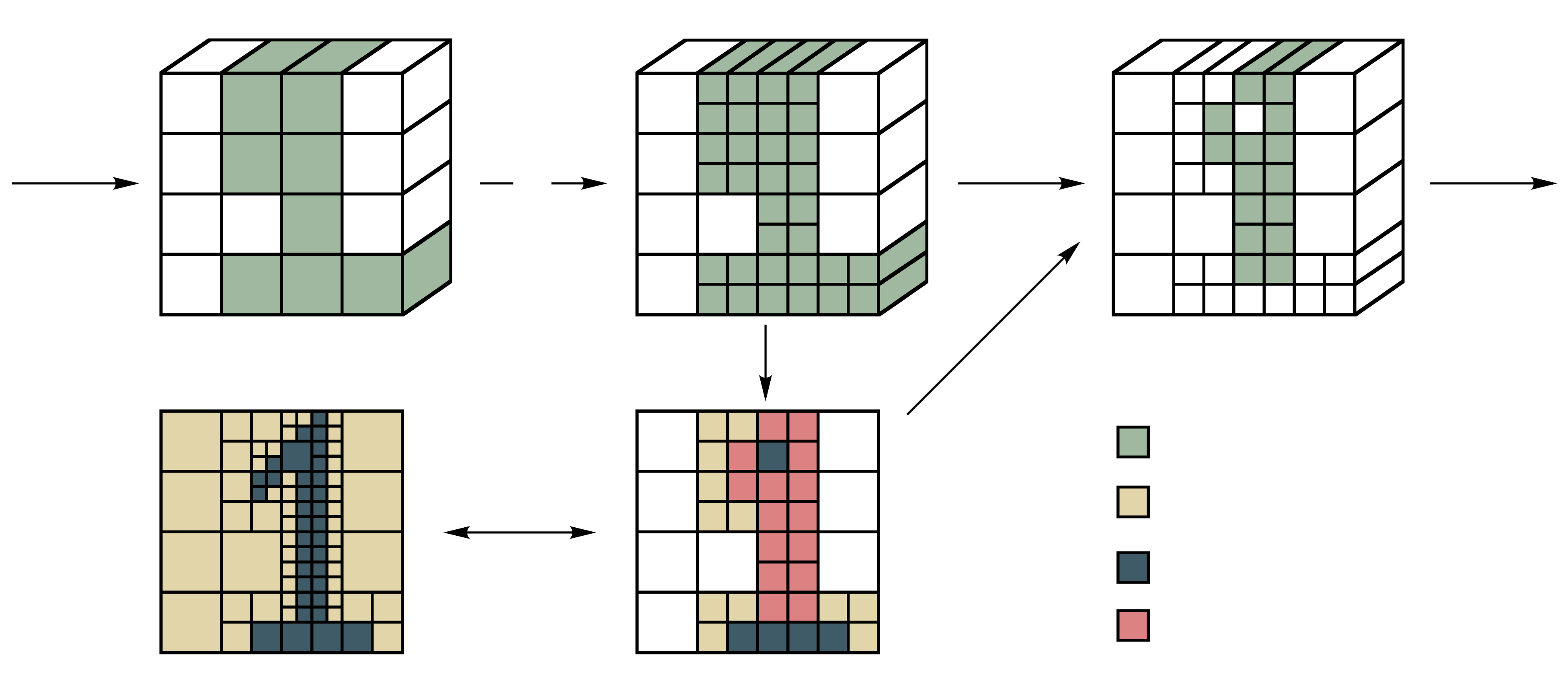}
 \put (3,33.5) {\huge ...}
 \put (93.5,33.5) {\huge ...}
 \put (30.2,33.5) {OGNConv}
 \put (29.3,29.7) {(one or more)}
 \put (33.25,32.1) {...}
 \put (27,23.5) {$c$}
 \put (18.5,22) {$d_2$}
 \put (8,30) {$d_1$}
 \put (60.9,33.5) {OGNProp}
 \put (28.9,11.5) {OGNLoss}
 \put (74,15.2) {propagated features}
 \put (74,11.4) {empty}
 \put (74,7.2) {filled}
 \put (74,3.5) {mixed}
 \put (19,43) {\large $F_{l-1}$}
 \put (49,43) {\large $\bar{F}_l$}
 \put (78.5,43) {\large $F_l$}
 \put (13,0) {Ground truth}
 \put (45,0) {Prediction}
 \put (51.5,21) {$1^3$}
 \put (50.1,19) {conv}
\end{overpic}
\end{center}
   \caption{Single block of an OGN illustrated as 2D quadtree for simplicity. After convolving features $F_{l-1}$ of the previous block with weight filters, we directly predict the occupancy values of cells at level $l$ using $1^3$ convolutions. Features corresponding to "filled" and "empty" cells are no longer needed and thus not propagated, which yields $F_l$ as the final output of this block.}
\label{fig:main}
\end{figure*}
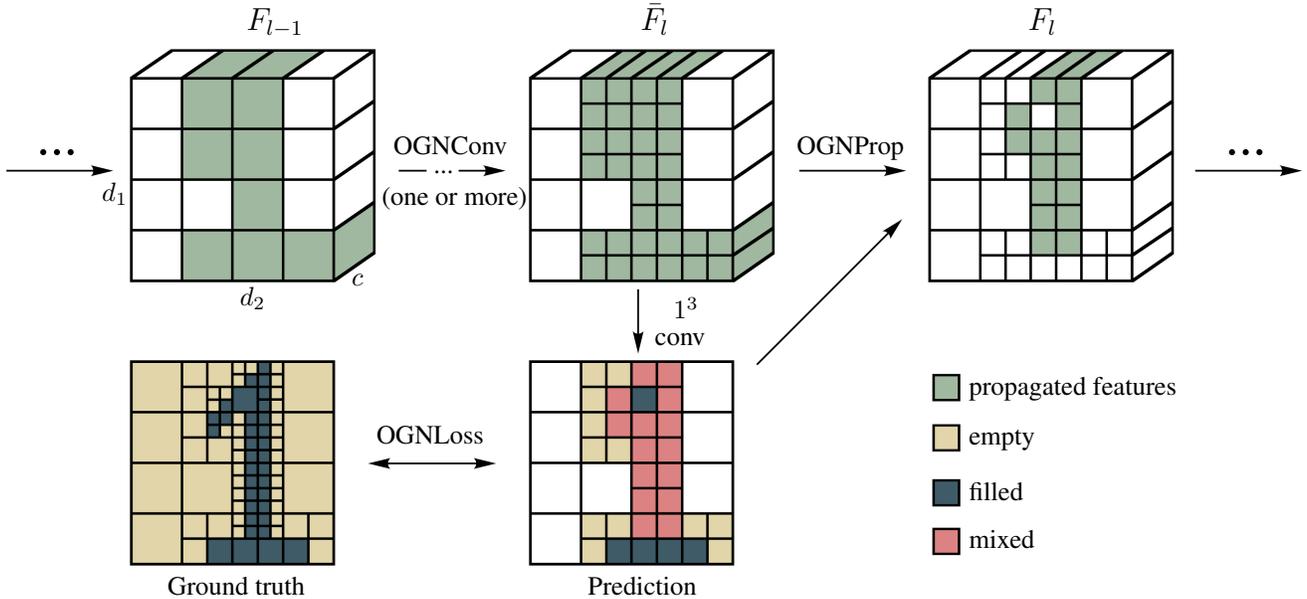

\section{Octrees}
\label{sec:octrees}

An octree \cite{octrees80} is a 3D grid structure with adaptive cell size, which allows for lossless reduction of memory consumption compared to a regular voxel grid.
\tcn{Octrees have a long history in classical 3D reconstruction and depth map fusion \cite{poisson, ssd11, depth_fusion, ummenhof, connolly, steinbrucker}.}
A function defined on a voxel grid can be converted into a function defined on an octree.
This can be done by starting from a single cell representing the entire space and recursively partitioning cells into eight octants.
If every voxel within a cell has the same function value, this cell is not subdivided and becomes a leaf of the tree.
The set of cells at a certain resolution is referred to as an octree level.
The recursive subdivision process can also be started not from the whole volume, but from some initial coarse resolution.
Then the maximal octree cell size is given by this initial resolution.
The most straightforward way of implementing an octree is to store in each cell pointers to its children.
In this case, the time to access an element scales linearly with the tree's depth, which can become costly at high resolutions.
We use a more efficient implementation that exploits hash tables.
An octree cell with spatial coordinates $\mathbf{x}=(x,y,z)$ at level $l$ is represented as an index-value pair $(m, v)$, where $v$ can be any kind of discrete or continuous signal.
$m$ is calculated from $(\mathbf{x},l)$ using Z-order curves \cite{octrees82}

\begin{equation}
m = \mathcal{Z}(\mathbf{x},l),
\end{equation}
which is a computationally cheap transformation implemented using bit shifts.
An octree $O$ is, hence, a set of all pairs
\begin{equation}
O=\{(m, v)\}.
\end{equation}
Storing this set as a hash table allows for constant-time element access.

When training networks, we will need to compare two different octrees $O^1$ and $O^2$, i.e. for each cell $(\mathbf{x}, l)$ from $O^1$, query the corresponding signal value $v$ in $O^2$.
Since different octrees have different structure, two situations are possible.
If $\mathcal{Z}(\mathbf{x},k)$  is stored at a level $k$ in $O^2$, which is the same or lower than $l$, the signal value of this cell can be uniquely determined.
If $\mathcal{Z}(\mathbf{x},k)$ is stored at one of the later levels, the cell is subdivided in $O^2$, and the value of the whole cell is not defined.
To formalize this, we introduce a function $f$ for querying the signal value of an arbitrary cell with index $m=\mathcal{Z}(\mathbf{x},l)$ from octree $O$:

\begin{equation}
\label{eq:state_func}
    f(m, O)=\left\{
                \begin{array}{ll}
                  v, \text{ if } \exists k \leq l: (\mathcal{Z}(\mathbf{x},k), v) \in O\\
                  \varnothing, \text{ otherwise}
                \end{array}
              \right.,
\end{equation}
where $\varnothing$ denotes an unavailable value.

\section{Octree Generating Networks}

An Octree Generating Network (\ognet{}) is a convolutional decoder that yields an octree as output: both the structure, i.e. which cells should be subdivided, and the signal value of each cell.
In this work we concentrate on binary occupancy values $v \in \{0,1\}$, but the proposed framework can be easily extended to support arbitrary signals.
As shown in Figure~\ref{fig:teaser}, an \ognet{} consists of a block operating on dense regular grids, followed by an arbitrary number of hash-table-based octree blocks.

The dense block is a set of conventional 3D convolutional and up-convolutional layers producing a feature map of size $d_1 \times d_2 \times d_3 \times c$ as output, where $\{d_i\}$ are the spatial dimension and $c$ is the number of channels.

From here on, the representation is processed by our custom layers operating on octrees.
The regular-grid-based feature map produced by the dense block is converted to a set of index-value pairs stored as a hash table (with values being feature vectors), and is further processed in this format. 
We organize octree-based layers in blocks, each responsible for predicting the structure and the content of a single level of the generated octree.

Figure~\ref{fig:main} illustrates the functioning of a single such block that predicts level $l$ of an octree.
For the sake of illustration, we replaced three-dimensional octrees by two-dimensional quadtrees.
Feature maps in Figure~\ref{fig:main} are shown as dense arrays only for simplicity; in fact the green cells are stored in hash maps, and the white cells are not stored at all.
We now give a high-level overview of the block and then describe its components in more detail.

Input to the block is a sparse hash-table-based convolutional feature map $F_{l-1}$ of resolution $(d_1 \cdot 2^{l-1}, d_2 \cdot 2^{l-1}, d_3 \cdot 2^{l-1})$ produced by the previous block.
First this feature map is processed with a series of custom convolutional layers and one up-convolutional layer with stride 2, all followed by non-linearities.

This yields a new feature map $\bar{F}_l$ of resolution $(d_1 \cdot 2^{l}, d_2 \cdot 2^{l}, d_3 \cdot 2^{l})$. 
Based on this feature map, we directly predict the content of level $l$.
For each cell, there is a two-fold decision to be made: should it be kept at level $l$, and if yes, what should be the signal value in this cell?
In our case making this decision can be formulated as classifying the cell as being in one of three states: "empty", "filled" or "mixed".
These states correspond to the outputs of state-querying function $f$ from eq.~\eqref{eq:state_func}, with "empty" and "filled" being the signal values $v$, and "mixed" being the state where the value is not determined.
We make this prediction using a convolutional layer with $1^3$ filters followed by a three-way softmax.
This classifier is trained in a supervised manner with targets provided by the ground truth octree.

Finally, in case the output resolution has not been reached, features from $\bar{F}_l$ that correspond to "mixed" cells are propagated to the next layer\footnote{Additional neighboring cells may have to be propagated if needed by subsequent convolutional layers. This is described in section~\ref{sec:propagation}.} and serve as an input feature map $F_l$ to the next block.

In the following subsections, we describe the components of a single octree block in more detail: the octree-based convolution, the loss function, and the feature propagation mechanism.

\subsection{Convolution}
\label{sec:convolution}
We implemented a custom convolutional layer \textit{OGNConv}, which operates on feature maps represented as hash tables instead of usual dense arrays.
Our implementation supports strided convolutions and up-convolutions with arbitrary filter sizes. It is based on representing convolution as a single matrix multiplication, similar to standard \textit{caffe}~\cite{caffe} code for dense convolutions.

\tcn{In the dense case, the feature tensor is converted to a matrix with the \textit{im2col} operation, then multiplied with the weight matrix of the layer, and the result is converted back into a dense feature tensor using the \textit{col2im} operation.
In OGN, instead of storing full dense feature tensors, only a sparse set of relevant features is stored at each layer.
These features are stored in a hash table, and we implemented custom operations to convert a hash table to a feature matrix and back.
The resulting matrices are much smaller than those in the dense case.
Convolution then amounts to multiplying the feature matrix by the weight matrix.
Matrix multiplication is executed on GPU with standard optimized functions, and our conversion routines currently run on CPU. 
Even with this suboptimal CPU implementation, computation times are comparable to those of usual dense convolutions at $32^3$ voxel resolution. At higher resolutions, \textit{OGNConv} is much faster than dense convolutions (see section~\ref{sec:speed}).}

Unlike convolutions on regular grids, OGN convolutions are not shift invariant by design. This is studied in Section \tcb{E} of the Appendix.

\subsection{Loss}
\label{sec:loss}
The classifier at level $l$ of the octree outputs the probabilities of each cell from this level being "empty", "filled" or "mixed", that is, a three-component prediction vector $\mathbf{p_m}=(p^0_m, p^1_m, p^2_m)$ for cell with index $m$.
We minimize the cross-entropy between the network predictions and the cell states of the ground truth octree $O_{GT}$, averaged over the set $M_l$ of cells predicted at layer $l$:

\begin{equation}
    \mathcal{L}_l = \frac{1}{|M_l|} \sum\limits_{m \in M_l}\left[\sum    \limits_{i=0}^2 h^i(f(m, O_{GT}))\log p^i_m\right],
  \end{equation}
where function $\mathbf{h}$ yields a one-hot encoding $(h^0, h^1, h^2)$ of the cell state value returned by $f$ from eq.~(\ref{eq:state_func}).
Loss computations are encapsulated in our custom \textit{OGNLoss} layer.

The final OGN objective is calculated as a sum of loss values from all octree levels
\begin{equation}
\mathcal{L} = \sum \limits_{l=1}^L \mathcal{L}_l.
\end{equation}

\subsection{Feature propagation}
\label{sec:propagation}

At the end of each octree block there is an \textit{OGNProp} layer that propagates to the next octree block features from cells in the "mixed" state, as well as from neighboring cells if needed to compute subsequent convolutions.
Information about the cell state can either be taken from the ground truth octree, or from the network prediction.
This spawns two possible propagation modes: using the known tree structure (\textit{Prop-known}) and using the predicted tree structure (\textit{Prop-pred}).
Section~\ref{sec:training} describes use cases for these two modes.

The set of features to be propagated 
depends on the kernel size in subsequent \textit{OGNConv} layers.
The example illustrated in Figure~\ref{fig:main} only holds for $2^3$ up-convolutions which do not require any neighboring elements to be computed.
To use larger convolutional filters or multiple convolutional layers, we must propagate not only the features of the "mixed" cells, but also the features of the neighboring cells required for computing the convolution at the locations of the "mixed" cells.
The size of the required neighborhood is computed based on the network architecture, before the training starts.
Details are provided in Section \tcb{C} of the Appendix.

\subsection{Training and testing}
\label{sec:training}

The OGN decoder is end-to-end trainable using standard backpropagation. The only subtlety is in selecting the feature propagation modes during training and testing. At training time the octree structure of the training samples is always available, and therefore the \textit{Prop-known} mode can be used. 
At test time, the octree structure may or may not be available. We have developed two training regimes for these two cases.

If the tree structure is available at test time, we simply train the network with \textit{Prop-known} and test it the same way. This regime is applicable for tasks like semantic segmentation, or, more generally, per-voxel prediction tasks, where the structure of the output is exactly the same as the structure of the input.

If the tree structure is not available at test time, we start by training the network with \textit{Prop-known}, and then fine-tune it with \textit{Prop-pred}. 
This regime is applicable to any task with volumetric output.

We have also tested other regimes of combining \textit{Prop-pred} and \textit{Prop-known} and found those to perform worse than the two described variants. This is discussed in more detail in Section \tcb{B} of the Appendix.

\section{Experiments}

In our experiments we verified that the \ognet{} architecture 
performs on par with the standard dense voxel grid representation, while requiring significantly less memory and computation, particularly at high resolutions. 
The focus of the experiments is on showcasing the capabilities of the proposed architecture. 
How to fully exploit the new architecture in practical applications is a separate problem that is left to future work.

\subsection{Experimental setup}
\label{sec:setup}

For all \ognet{} decoders used in our evaluations, we followed the same design pattern: 1 or 2 up-convolutional layers interleaved with a convolutional layer in the dense block, followed by multiple octree blocks depending on the output resolution.
In the octree blocks we used $2^3$ up-convolutions.
We also evaluated two other architecture variants, presented in section~\ref{sec:architecture}.
ReLU non-linearities were applied after each (up-)convolutional layer.
The number of channels in the up-convolutional layers of the octree blocks was set to 32 in the outermost layer, and was increased by 16 in each preceding octree block.
The exact network architectures used in individual experiments are shown in Section \tcb{F} of the Appendix.

The networks were trained using ADAM \cite{adam} with initial learning rate 0.001, $\beta_1=0.9$, $\beta_2=0.999$.
The learning rate was decreased by a factor of 10 after 30K and 70K iterations.
We did not apply any additional regularization.

For quantitative evaluations, we converted the resulting octrees back to regular voxel grids, and computed the Intersection over Union (IoU) measure between the ground truth model and the predicted model.
To quantify the importance of high-resolution representations, in some experiments we upsampled low-resolution network predictions to high-resolution ground truth using trilinear interpolation\tcn{, and later binarization with a threshold of 0.5.}
We explicitly specify the ground truth resolution in all experiments where this was done.

\tcn{If not indicated otherwise, the results were obtained in the \textit{Prop-pred} mode.}

\subsubsection{Datasets}
In our evaluations we used three datasets:

\textbf{\shapenetall{}} 
Approximately 50.000 CAD models from 13 main categories of the ShapeNet dataset \cite{shapenet2015}, used by Choy et al. \cite{choy20163d}.
We also used the renderings provided by Choy et al.~\cite{choy20163d}.

\textbf{\shapenetcars{}} 
A subset of \shapenetall{} consisting of 7497 car models.

\textbf{\blendswap{}} 
A dataset of 4 scenes we manually collected from \url{blendswap.com}\tcn{, a website containing a large collection of Blender models.}

\tcn{All datasets were voxelized in multiple resolutions from $32^3$ to $512^3$ using the \textit{binvox}\footnote{\url{http://www.patrickmin.com/binvox}} tool, and were converted into octrees.}
We set the interior parts of individual objects to be filled, and the exterior to be empty.

\subsection{Computational efficiency}
\label{sec:speed}

We start by empirically demonstrating that \ognet{}s can be used at high resolutions when the voxel grid representation becomes impractical both because of the memory requirements and the runtime.

The number of elements in a voxel grid is uniquely determined by its resolution, and scales cubically as the latter increases.
The number of elements in an octree depends on the data, leading to variable scaling rates: from constant for cubic objects aligned with the grid, to cubic for pathological shapes such as a three-dimensional checkerboard.
In practice, octrees corresponding to real-world objects and scenes scale approximately quadratically, since they represent smooth two-dimensional surfaces in a three-dimensional space.

\begin{figure}[hb]
\centering
\begin{minipage}{.5\linewidth}
\begin{tikzpicture}
	\begin{axis}[
	    height=4cm,
	    width=5.1cm,
	    legend pos=north west,
	    xtick={32, 128, 256, 512},
		ylabel near ticks,
        xlabel near ticks,
        style={font=\scriptsize}]
	\addplot [color=red, mark=square*, mark size=1.2] coordinates {
	    (32,0.29)
	    (64,0.36)
		(128,0.43)
		(256,0.54)
		(512,0.88)
	};
 	\addplot [color=blue, mark=*, mark size=1.2] coordinates {
 	    (32,0.33)
 	    (64,0.5)
 		(128,1.62)
 		(256,9.98)
 	};
 	\addplot [dashed, color=blue, mark=*, mark size=1.2] coordinates {
 		(256,9.98)
 		(512,74.28)
 	};
    \legend{\ognet{}, Dense}
	\end{axis}
	
\node at (1.7cm,2.7cm) {\scriptsize Memory, GB};

\end{tikzpicture}
\end{minipage}%
\begin{minipage}{.5\linewidth}
\begin{tikzpicture}
	\begin{axis}[
	    height=4cm,
	    width=5.1cm,
	    legend pos=north west,
	    xtick={32, 128, 256, 512},
        xlabel near ticks,
        ylabel near ticks,
        style={font=\scriptsize}]
	\addplot [color=red, mark=square*, mark size=1.2] coordinates {
	    (32,0.016)
	    (64,0.057)
		(128,0.183)
		(256,0.641)
		(512,2.056)
	};
 	\addplot [color=blue, mark=*, mark size=1.2] coordinates {
 	    (32,0.015)
 	    (64,0.193)
 		(128,0.561)
 		(256,3.891)
 	};
 	\addplot [dashed, color=blue, mark=*, mark size=1.2] coordinates {
 	    (256,3.891)
 		(512,41.295)
 	};
    \legend{\ognet{}, Dense}
	\end{axis}

\node at (1.7cm,2.7cm) {\scriptsize Iteration time, s};

\end{tikzpicture}
\end{minipage}
\vspace{0.1cm} 
\caption{Memory consumption (left) and iteration time (right) of OGN and a dense network at different output resolutions. Forward and backward pass, batch size 1.}
\label{fig:efficiency}
\end{figure}
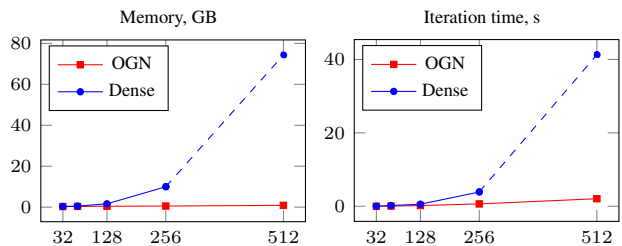

\begin{table}[thb]
\begin{center}
\begin{tabular}{|c|c|c|c|c|}
\hline
        & \multicolumn{2}{c|}{Memory, GB} & \multicolumn{2}{c|}{Iteration time, s} \\
\hline
Resolution & Dense & OGN & Dense & OGN \\
\hline\hline
$32^3$ & 0.33 & \textbf{0.29} & \textbf{0.015} & 0.016 \\
$64^3$ & 0.50 & \textbf{0.36} & 0.19 & \textbf{0.06} \\
$128^3$ & 1.62 & \textbf{0.43} & 0.56 & \textbf{0.18} \\
$256^3$ & 9.98 & \textbf{0.54} & 3.89 & \textbf{0.64} \\
$512^3$ & (74.28) & \textbf{0.88} & (41.3) & \textbf{2.06} \\
\hline
\end{tabular}
\end{center} 
\caption{Memory consumption and iteration time of OGN and a dense network at different output resolutions. Batch size 1.}
\label{tbl:runtime}
\end{table}

We empirically compare the runtime and memory consumption values for a dense network and \ognet{}, for varying output resolution.
Architectures of the networks are the same as used in Section~\ref{sec:decoder}~-- three fully connected layers followed by an up-convolutional decoder. 
We performed the measurements on an NVidia TitanX Maxwell GPU, with 12Gb of memory.
To provide actual measurements for dense networks at the largest possible resolution, we performed the comparison with batch size 1.
The $512^3$ dense network does not fit into memory even with batch size 1, so we extrapolated the numbers by fitting cubic curves.

Figure~\ref{fig:efficiency} and Table~\ref{tbl:runtime} show the results of the comparison.
The OGN is roughly as efficient as its dense counterpart for low resolutions, but as the resolution grows, it gets drastically faster and consumes far less memory.
At $512^3$ voxel resolution, the \ognet{} consumes almost two orders of magnitude less memory and runs $20$ times faster.
In Section \tcb{A} of the Appendix we provide a more detailed analysis and explicitly demonstrate the near-cubic scaling of dense networks against the near-quadratic scaling of OGNs.

To put these numbers into perspective, training \ognet{} at $256^3$ voxel output resolution takes approximately $5$ days.
Estimated training time of its dense counterpart would be almost a month.
Even if the $512^3$ voxel dense network would fit into memory, it would take many months to train.

\subsection{Autoencoders}

Autoencoders and their variants are commonly used for representation learning from volumetric data ~\cite{Girdhar16b, SharmaGF16}.
Therefore, we start by comparing the representational power of the \ognet{} to that of dense voxel grid networks on the task of auto-encoding volumetric shapes.

We used the decoder architecture described in section~\ref{sec:setup} both for the \ognet{} and the dense baseline.
The architecture of the encoder is symmetric to the decoder.
Both encoders operate on a dense voxel grid representation\footnote{In this paper, we focus on generating 3D shapes. Thus, we have not implemented an octree-based convolutional encoder. This could be done along the lines of Riegler et al.~\cite{octnet}}.

We trained the autoencoders on the \shapenetcars{} dataset in two resolutions: $32^3$ and $64^3$.
\tcn{We used 80$\%$ of the data for training, and 20$\%$ for testing.}
Quantitative results are summarized in Table~\ref{tbl:autoencoder}.
With predicted octree structure, there is no significant difference in performance between the \ognet{} and the dense baseline.

\begin{table}[h]
\begin{center}
\begin{tabular}{|l|c|c|}
\hline
Network & $32^3$ & $64^3$ \\
\hline\hline
Dense & 0.924 & 0.890 \\
\ognet{}+\textit{Prop-known} & 0.939 & 0.904 \\
\ognet{}+\textit{Prop-pred} & 0.924 & 0.884 \\
\hline
\end{tabular}
\end{center} 
\caption{Quantitative results for OGN and dense autoencoders. Predictions were compared with the ground truth at the corresponding resolution, without upsampling.}
\label{tbl:autoencoder}
\end{table}
\subsubsection{Flexibility of architecture choice}
\label{sec:architecture}

To show that OGNs are not limited to up-convolutional layers with $2^3$ filters, we 
evaluated two alternative $64^3$ \ognet{} auto-encoders: one with $4^3$ up-convolutions and one with $2^3$ up-convolutions interleaved with $3^3$ convolutions.
The results are summarized in Table~\ref{tbl:technical}.
There is no significant difference between the architectures for this task. 
With larger filters, the network is roughly twice slower in our current implementation, so we used $2^3$ filters in all further experiments.

\begin{table}
\begin{center}
\begin{tabular}{|l|c|c|c|c|}
\hline
Mode & 2x2 filters & 4x4 filters & IntConv \\
\hline\hline
\ognet{}+\textit{Prop-known} & 0.904 & 0.907 & 0.907 \\
\ognet{}+\textit{Prop-pred} & 0.884 & 0.885 & 0.885 \\
\hline
\end{tabular}
\end{center}
\caption{Using more complex architectures in $64^3$ OGN autoencoders does not lead to significant performance improvements.}
\label{tbl:technical}
\end{table}

\subsubsection{Using known structure}
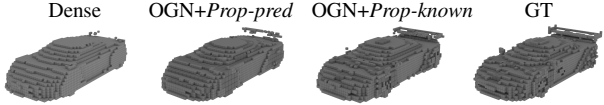
\begin{figure}[ht]
\begin{center}
\vspace{0.2cm}
\begin{overpic}[width=\linewidth]{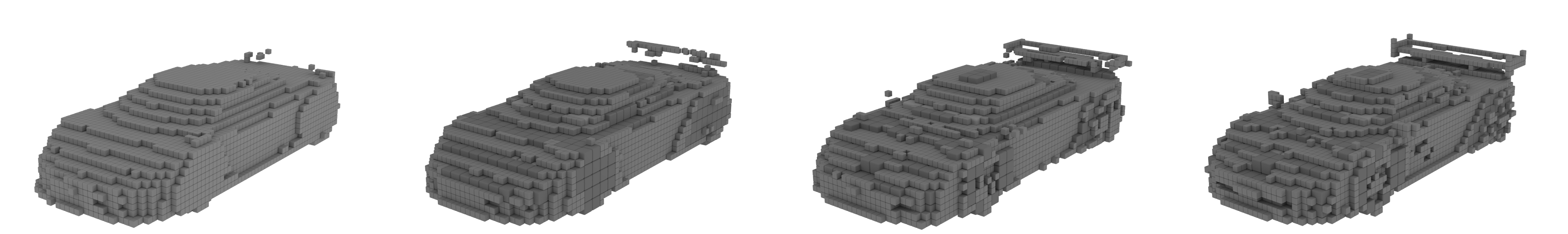}
 \put (9,14) {\footnotesize Dense}
 \put (25,14) {\footnotesize \ognet{}+\textit{Prop-pred}}
 \put (51,14) {\footnotesize \ognet{}+\textit{Prop-known}}
 \put (85,14) {\footnotesize GT}
\end{overpic}
\end{center}
   \caption{Using the known tree structure at test time leads to improved performance.}
\label{fig:autoencoder_comparison}
\end{figure}

Interestingly, \ognet{} with known tree structure outperforms the network based on a dense voxel grid, both qualitatively and quantitatively.
An example of this effect can be seen in Figure~\ref{fig:autoencoder_comparison}: the dense autoencoder and our autoencoder with predicted propagation struggle with properly reconstructing the spoiler of the car.
Intuitively, the known tree structure provides additional shape information to the decoder, thus simplifying the learning problem.
In the autoencoder scenario, however, this may be undesirable if one aims to encode all information about a shape in a latent vector.
In tasks like semantic segmentation, the input octree structure could help introduce shape features implicitly in the learning task.

\subsection{3D shape from high-level information}
\label{sec:decoder}

We trained multiple \ognet{}s for generating shapes from high-level parameters similar to Dosovitskiy et al.~\cite{chairs_pami}.
In all cases the input of the network is a one-hot encoded object ID, and the output is an octree with the object shape.

\subsubsection{\shapenetcars{}}
\label{sec:decoder_cars}
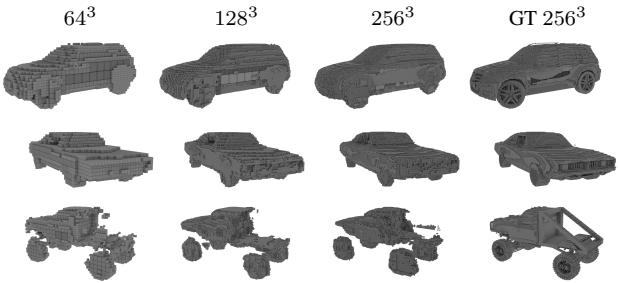
\begin{figure}[t]
\begin{center}
\begin{overpic}[width=\linewidth]{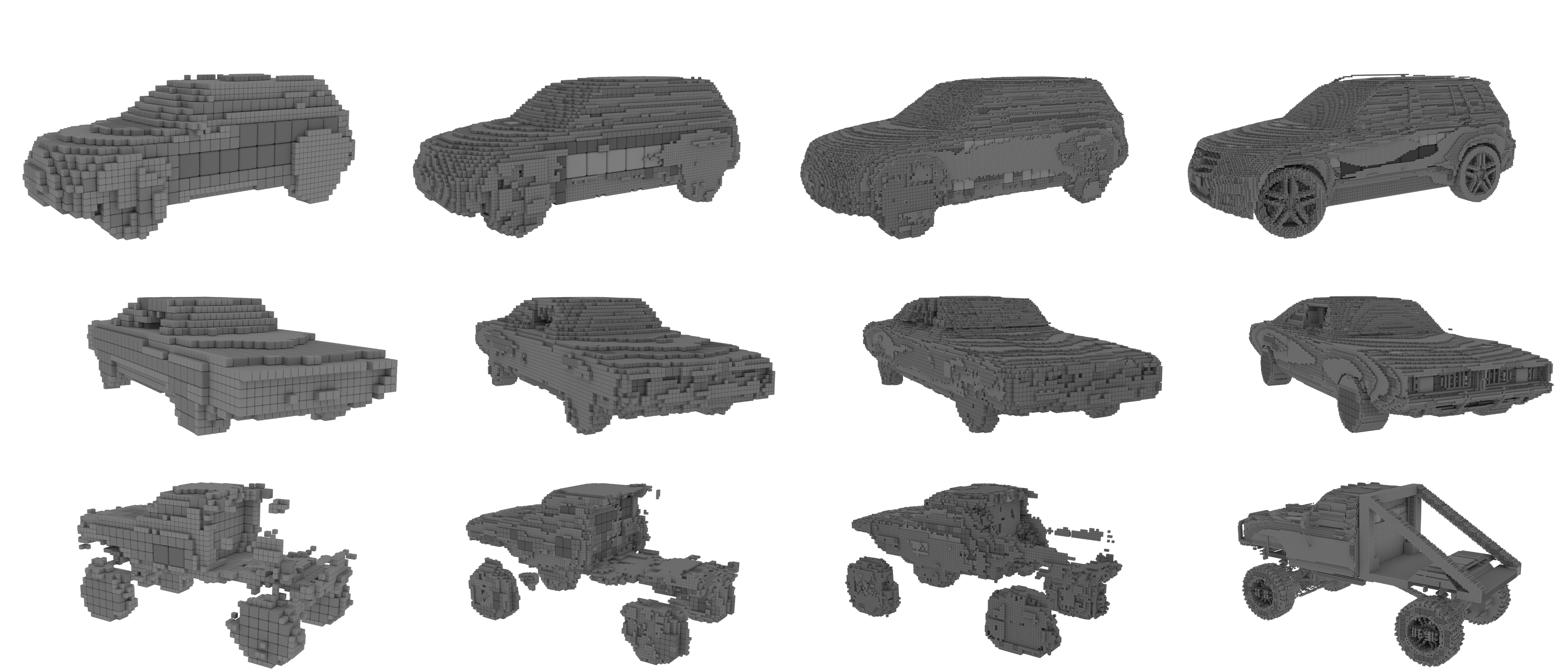}
 \put (11,41) {\footnotesize $64^3$}
 \put (35,41) {\footnotesize $128^3$}
 \put (60,41) {\footnotesize $256^3$}
 \put (82,41) {\footnotesize GT $256^3$}
\end{overpic}
\end{center}
   \caption{\tcn{Training} samples from the \shapenetcars{} dataset generated by our networks. Cells at different octree levels vary in size and are displayed in different shades of gray.}
\label{fig:decoder_example}
\end{figure}

First, we trained on the whole \shapenetcars{} dataset in three resolutions: $64^3$, $128^3$ and $256^3$.
Example outputs are shown in Figure~\ref{fig:decoder_example} and quantitative results are presented in Table~\ref{tbl:across_resolutions}.
Similar to the two-dimensional case \cite{chairs_pami}, the outputs are accurate in the overall shape, but lack some fine details.
This is not due to the missing resolution, but due to general limitations of the training data and the learning task. 
Table~\ref{tbl:across_resolutions} reveals that a resolution of $128^3$ allows the reconstruction of a more accurate shape with more details than a resolution of $64^3$. At an even higher resolution of $256^3$, the overall performance decreased again. 
Even though the higher-resolution network is architecturally capable of performing better, it is not guaranteed to train better.
Noisy gradients from outer high-resolution layers may hamper learning of deeper layers, resulting in an overall decline in performance.
This problem is orthogonal to the issue of designing computationally efficient architectures, which we aim to solve in this paper. 
We further discuss this in the Appendix.

\begin{table}
\begin{center}
\begin{tabular}{|l|c|c|c|c|}
\hline
         & 64 & 128 & 256 & 512\\
\hline
\shapenetcars{} & 0.856 & \textbf{0.901} & 0.865 & - \\
\blendswap{} & 0.535 & 0.649 & 0.880 & \textbf{0.969} \\
\hline
\end{tabular}
\end{center}
\caption{Quantitative evaluation of 3D shapes generated from high-level information. Lower-resolution predictions from \shapenetcars{} were upsampled to $256^3$ ground truth, scenes from \blendswap{}~--- to $512^3$.}
\label{tbl:across_resolutions}
\end{table}

\begin{figure}[h]
\begin{center}
\begin{overpic}[width=\linewidth]{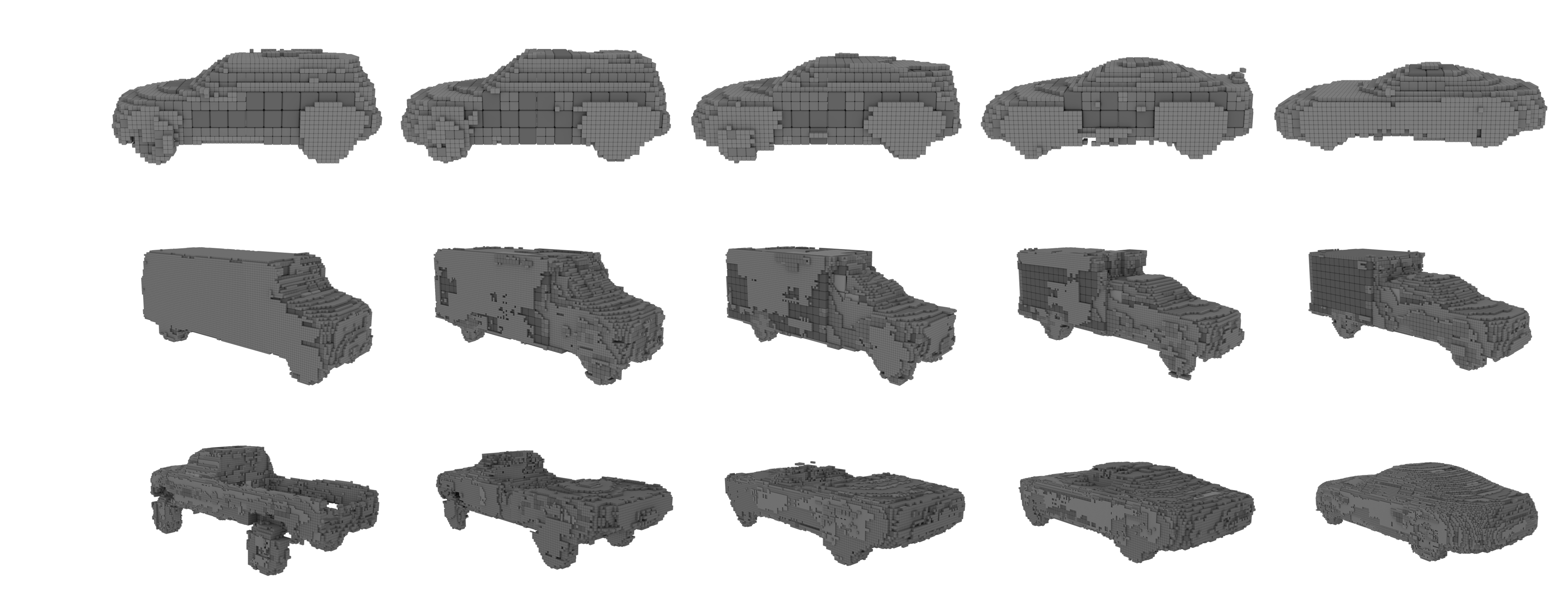}
 \put (0,30) {\footnotesize $64^3$}
 \put (0,17) {\footnotesize $128^3$}
 \put (0,4) {\footnotesize $256^3$}
\end{overpic}
\end{center}
   \caption{Our networks can generate previously unseen cars by interpolating between the dataset points, which demonstrates their generalization capabilities.}
\label{fig:morphing}
\end{figure}

Notably, the network does not only learn to generate objects from the training dataset, but it can also generalize to unseen models.
We demonstrate this by interpolating between pairs of one-hot input ID vectors.
Figure~\ref{fig:morphing} shows that for all intermediate input values the network produces consistent output cars, with the style being smoothly changed between the two training points.

\subsubsection{\blendswap{}}

\tcn{To additionally showcase the benefit of using higher resolutions, we trained \ognet{}s to fit the \blendswap{} dataset containing 4 whole scenes.
In contrast to the \shapenetcars{} dataset, such amount of training data does not allow for any generalization.
The experiment aims to show that \ognet{}s provide sufficient resolution to represent such high-fidelity shape data.}

\begin{figure*}[t]
\begin{center}
\begin{overpic}[width=\linewidth]{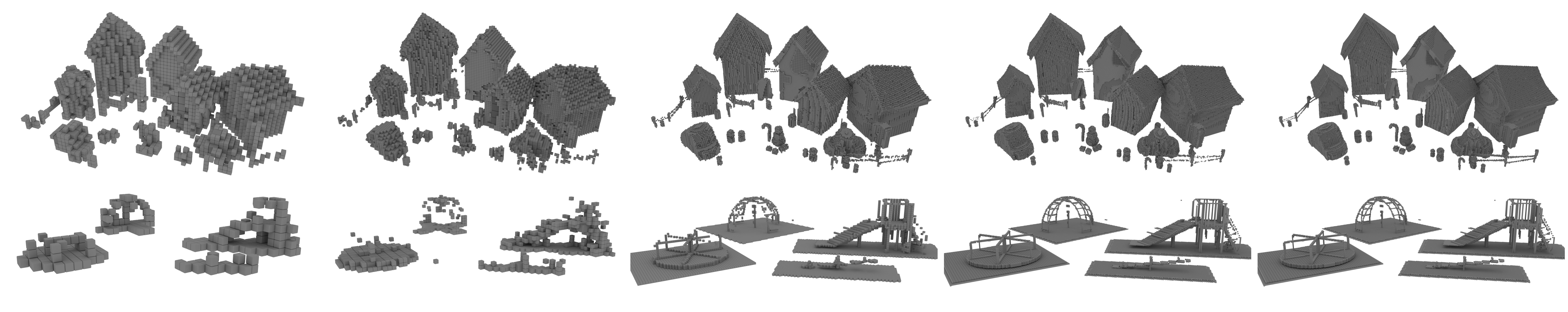}
 \put (9,20.5) {\footnotesize $64^3$}
 \put (28,20.5) {\footnotesize $128^3$}
 \put (48,20.5) {\footnotesize $256^3$}
 \put (68,20.5) {\footnotesize $512^3$}
 \put (86.5,20.5) {\footnotesize GT $512^3$}
\end{overpic}
\end{center}
   \caption{OGN is used to \tcn{reproduce} large-scale scenes \tcn{from the dataset}, where high resolution is crucial to generate fine-grained structures.}
\label{fig:scenes}
\end{figure*}
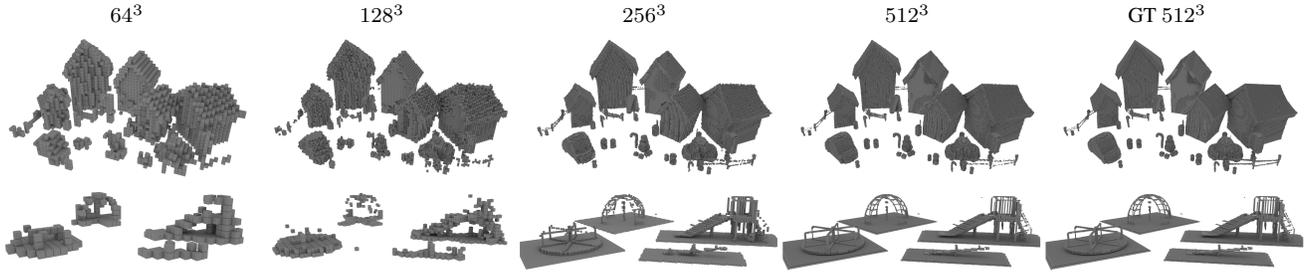

\tcn{Figure~\ref{fig:scenes} shows the generated scenes.
In both examples, $64^3$ and $128^3$ resolutions are inadequate for representing the details. 
For the bottom scene, even the $256^3$ resolution still struggles with fine-grained details.
This example demonstrates that tasks like end-to-end learning of scene reconstruction requires high-resolution representations, and the \ognet{} is an architecture that can provide such resolutions.}

\tcn{These qualitative observations 
are confirmed quantitatively in Table~\ref{tbl:across_resolutions}. Higher output resolutions allow for more accurate  reconstruction of the samples in the dataset.}
More results for this experiment are shown in Section \tcb{D} of the Appendix, and the accompanying video\footnote{\url{https://youtu.be/kmMvKNNyYF4}}.

\subsection{Single-image 3D reconstruction}

In this experiment we trained networks with our \ognet{} decoder on the task of single-view 3D reconstruction. 
To demonstrate that our dense voxel grid baseline, as already used in the autoencoder experiment, is a strong baseline, we compare to the approach by Choy et al. \cite{choy20163d}.
This approach operates on $32^3$ voxel grids, and we adopt this resolution for our first experiment. 
To ensure a fair comparison, we trained networks on \shapenetall{}, the exact dataset used by Choy et al.~\cite{choy20163d}.
\tcn{Following the same dataset splitting strategy, we used 80$\%$ of the data for training, and 20$\%$ for testing.}
As a baseline, we trained a network with a dense decoder which had the same configuration as our OGN decoder.
Table~\ref{tbl:2d-3d_full} shows that compared to single-view reconstructions from \cite{choy20163d}, both the OGN and the baseline dense network compare favorably for most of the classes.
In conclusion, the \ognet{} is competitive with voxel-grid-based networks on the complex task of single-image class-specific 3D reconstruction.

\begin{table}
\begin{center}
\begin{tabular}{|l|c|c|c|}
\hline
Category & R2N2 \cite{choy20163d} & \ognet{} & Dense \\
\hline\hline
Plane & 0.513 & \textbf{0.587} & 0.570 \\
Bench & 0.421 & \textbf{0.481} & 0.481 \\
Cabinet & 0.716 & 0.729 & \textbf{0.747} \\
Car & 0.798 & 0.816 & \textbf{0.828} \\
Chair & 0.466 & \textbf{0.483} & 0.481 \\
Monitor & 0.468 & 0.502 & \textbf{0.509} \\
Lamp & 0.381 & \textbf{0.398} & 0.371 \\
Speaker & \textbf{0.662} & 0.637 & 0.650 \\
Firearm & 0.544 & \textbf{0.593} & 0.576 \\
Couch & 0.628 & 0.646 & \textbf{0.668} \\
Table & 0.513 & 0.536 & \textbf{0.545} \\
Cellphone & 0.661 & \textbf{0.702} & 0.698 \\
Watercraft & 0.513 & \textbf{0.632} & 0.550 \\
\hline\hline
Mean & 0.560 & \textbf{0.596} & 0.590 \\
\hline
\end{tabular}
\end{center}
\caption{Single-view 3D reconstruction results on the $32^3$ version of \shapenetall{} from Choy et al.~\cite{choy20163d} compared to OGN and a dense baseline. OGN is competitive with voxel-grid-based networks.}
\label{tbl:2d-3d_full}
\end{table}

\begin{figure}
\begin{center}
\begin{overpic}[width=\linewidth]{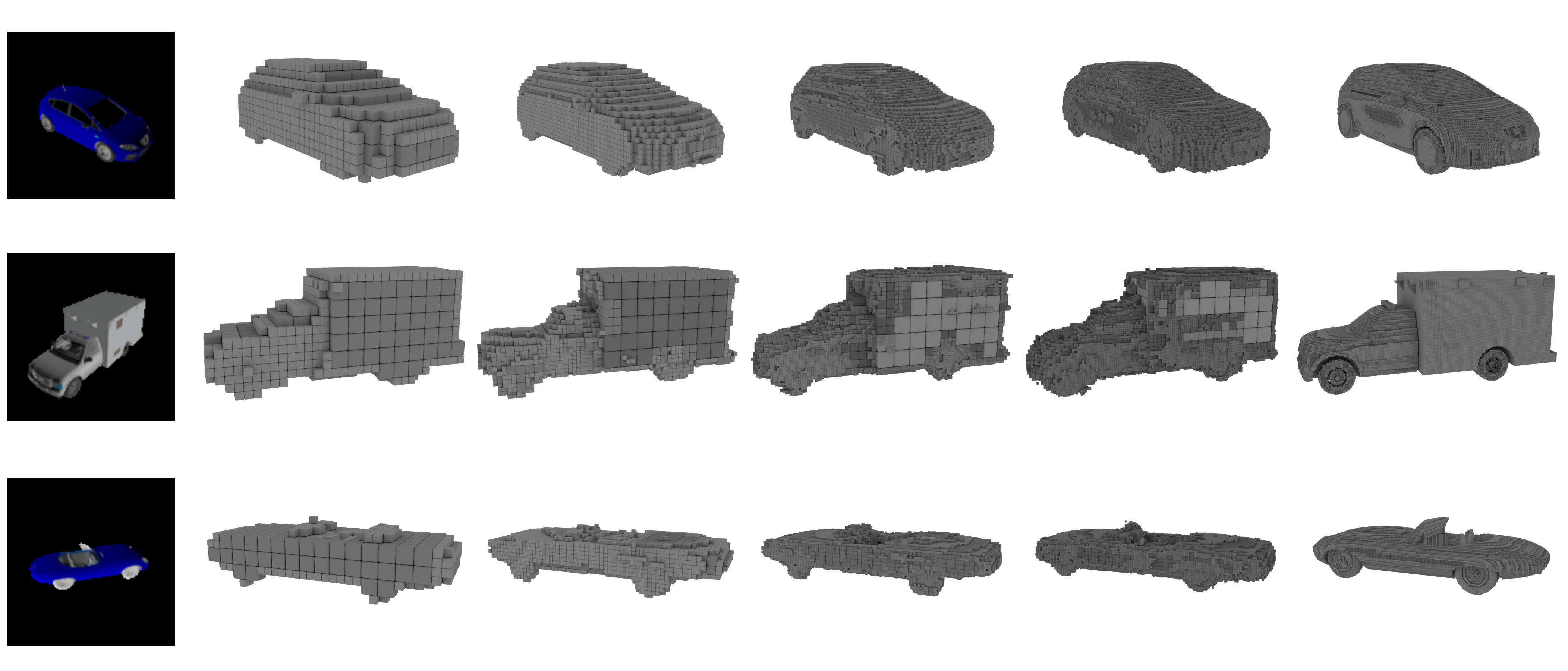}
 \put (2,42) {\footnotesize Input}
 \put (19,42) {\footnotesize $32^3$}
 \put (36,42) {\footnotesize $64^3$}
 \put (52.5,42) {\footnotesize $128^3$}
 \put (70,42) {\footnotesize $256^3$}
 \put (85,42) {\footnotesize GT $256^3$}
\end{overpic}
\end{center}
   \caption{Single-image 3D reconstruction on the \shapenetcars{} dataset using \ognet{} in different resolutions.}
\label{fig:2d-3d-all}
\end{figure}
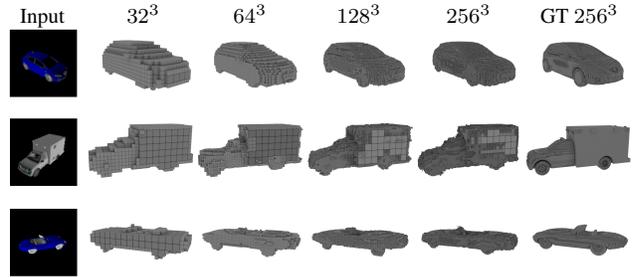

We also evaluated the effect of resolution on the \shapenetcars{} dataset.
Figure~\ref{fig:2d-3d-all} shows that \ognet{}s learned to infer the 3D shapes of cars in all cases, and that high-resolution predictions are clearly better than the $32^3$ models commonly used so far.
This is backed up by quantitative results shown in Table~\ref{tbl:iou_2d-3d}: $32^3$ results are significantly worse than the rest.
At $256^3$ performance drops again for the same reasons as in the decoder experiment in section~\ref{sec:decoder_cars}.

\begin{table}
\begin{center}
\begin{tabular}{|l|c|c|c|c|}
\hline
Resolution & 32 & 64 & 128 & 256 \\
\hline
Single-view 3D & 0.641 & 0.771 & \textbf{0.782} & 0.766 \\
\hline
\end{tabular}
\end{center}
\caption{Single-image 3D reconstruction results on \shapenetcars{}. Low-resolution predictions are upsampled to $256^3$. Commonly used $32^3$ models are significantly worse than the rest.}
\label{tbl:iou_2d-3d}
\end{table}

\section{Conclusions}

We have presented a novel convolutional decoder architecture for generating high-resolution 3D outputs represented as octrees. We have demonstrated that this architecture is flexible in terms of the exact layer configuration, and that it provides the same accuracy as dense voxel grids in low resolution. At the same time, it scales much better to higher resolutions, both in terms of memory and runtime.

This architecture enables end-to-end deep learning to be applied to tasks that appeared unfeasible before. In particular, learning tasks that involve 3D shapes, such as 3D object and scene reconstruction, are likely to benefit from it.

While in this paper we have focused on shapes and binary occupancy maps, it is straightforward to extend the framework to multi-dimensional outputs attached to the octree structure; for example, the output of the network could be a textured shape or a signed distance function. This will allow for an even wider range of applications.

\section*{Acknowledgements}

This work was supported by the Excellence Initiative of the German Federal and State Governments: BIOSS Centre for Biological Signalling Studies (EXC 294).
\tcn{We would like to thank Benjamin Ummenhofer for valuable discussions and technical comments.}
\tcn{We also thank Nikolaus Mayer for his help with 3D model visualization and manuscript preparation.}

\clearpage

{\small
\bibliographystyle{ieee}
\bibliography{egbib}
}

\clearpage

\section*{Appendix}

\appendix

\section{Computational efficiency}

In the main paper we have shown that with a practical architecture our networks scale much better than their dense counterparts both in terms of memory consumption and computation time.
The numbers were obtained for the "houses" scene from the BlendSwap dataset.

\begin{figure}[ht]
\begin{center}
\begin{tikzpicture}
	\begin{axis}[
	    width=\linewidth,
	    legend pos=north west,
	    xmode=log,
	    ymode=log,
	    xtick={32, 64, 128, 256, 512},
	    log ticks with fixed point,
		xlabel=Resolution,
		ylabel={Peak memory usage, GB}]
	\addplot [color=red, mark=square*] coordinates {
	    (32,4 / 1024)
		(64,7 / 1024)
		(128,12 / 1024)
		(256,24 / 1024)
		(512,81 / 1024)
	};
	\addplot [color=red, dashed, mark=x] coordinates {
		(32,4 / 1024)
		(64,16 / 1024)
		(128,64 / 1024)
		(256,256 / 1024)
		(512,1024 / 1024)
	};
	\addplot [color=blue, mark=*] coordinates {
	    (32,8 / 1024)
		(64,22 / 1024)
		(128,132 / 1024)
		(256,1004 / 1024)
		(512,7981 / 1024)
	};
	\addplot [color=blue, dashed, mark=x] coordinates {
	    (32,8 / 1024)
		(64,64 / 1024)
		(128,512 / 1024)
		(256,4096 / 1024)
		(512,32768 / 1024)
	};
    \legend{OGN,Quadratic,Dense,Cubic}
	\end{axis}

\end{tikzpicture}
\end{center}
\caption{Memory consumption for very slim networks, forward and backward pass, batch size 1. Shown in log-log scale - lines with smaller slope correspond to better scaling.}
\label{fig:memory_slim}
\end{figure}
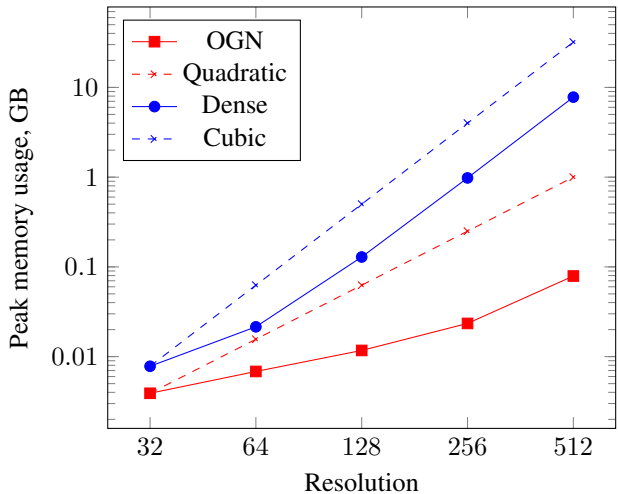

\begin{figure}[ht]
\begin{center}
\begin{tikzpicture}
	\begin{axis}[
	    width=\linewidth,
	    legend pos=north west,
	    xmode=log,
	    ymode=log,
	    xtick={32, 64, 128, 256, 512},
	    log ticks with fixed point,
		xlabel=Resolution,
		xlabel near ticks,
		ylabel={Iteration time, s}]
	\addplot [color=red, mark=square*, mark options={solid}] coordinates {
	    (32,0.011)
		(64,0.022)
		(128,0.083)
		(256,0.204)
		(512,0.975)
	};
	\addplot [color=red, dashed, mark=square, mark options={solid}] coordinates {
		(32,4 / 364)
		(64,16 / 364)
		(128,64 / 364)
		(256,256 / 364)
		(512,1024 / 364)
	};
	\addplot [color=blue, mark=*, mark options={solid}] coordinates {
	    (32,0.009)
		(64,0.033)
		(128,0.2)
		(256,1.349)
		(512,10.058)
	};
	\addplot [color=blue, dashed, mark=o, mark options={solid}] coordinates {
	    (32,8 / 889)
		(64,64 / 889)
		(128,512 / 889)
		(256,4096 / 889)
		(512,32768 / 889)
	};
    \legend{OGN,Quadratic,Dense,Cubic}
	\end{axis}

\end{tikzpicture}
\end{center}
\caption{Iteration time for very slim networks, forward and backward pass, batch size 1. Shown in log-log scale - lines with smaller slope correspond to better scaling.}
\label{fig:runtime_slim}
\end{figure}
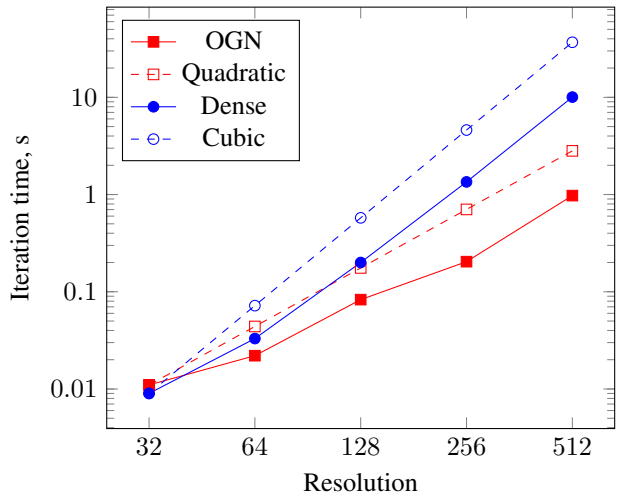

In order to further study this matter, we have designed a set of slim decoder networks that fit on a GPU in every resolution, including $512^3$, both with an OGN and a dense representation.
The architectures of those networks are similar to those from Table~\ref{tbl:architectures}, but with only 1 channel in every convolutional layer, and a single fully-connected layer with 64 units in the encoder.
The resulting measurements are shown in Figure~\ref{fig:memory_slim} for memory consumption and Figure~\ref{fig:runtime_slim} for runtime.
To precisely quantify the scaling, we subtracted the constant amount of memory reserved on a GPU by \textit{caffe} (190 MB) from all numbers.

Both plots are displayed in the log-log scale, i.e., functions from the family $y=a x^k$ are straight lines. 
The slope of this line is determined by the exponent $k$, and the vertical shift by the coefficient $a$.
In this experiment we are mainly interested in the slope, that is, how do the approaches scale with increasing output resolution.
As a reference, we show dashed lines corresponding to perfect cubic and perfect quadratic scaling.

Starting from $64^3$ voxel resolution both the runtime and the memory consumption scale almost cubically in case of dense networks.
For this particular example, OGN scales even better than quadratically, but in general scaling of the octree-based representation depends on the specific data it is applied to.

\section{Train/test modes}

In Section 4.4 of the main paper, we described how we use the two propagation modes (\textit{Prop-known} and \textit{Prop-pred}) during training and testing.
Here we motivate the proposed regimes, and show additional results with other combinations of propagation modes.

When the structure of the output tree is not known at test time, we train the networks until convergence with \textit{Prop-known}, and then additionally fine-tune with \textit{Prop-pred} - line 4 in Table~\ref{tbl:technical}.
Without this fine-tuning step (line 2), there is a decrease in performance, which is more significant when using larger convolutional filters.
Intuitively, this happens because the network has never seen erroneous propagations during training, and does not now how to deal with them at test time.

When the structure of the output is known at test time, the best strategy is to simply train in \textit{Prop-known}, and test the same way (line 1).
Additional fine-tuning in the \textit{Prop-pred} mode slightly hurts performance in this case (line 3).
The overall conclusion is not surprising: the best results are obtained when training networks in the same propagation modes, in which they are later tested.

\begin{table}
\begin{center}
\begin{tabular}{|c|c|c|c|c|c|}
\hline
Training & Testing & $2^3$ filters & $4^3$ filters & IntConv \\
\hline\hline
\textit{Known} & \textit{Known} & 0.904 & 0.907 & 0.907 \\
\textit{Known} & \textit{Pred} & 0.862 & 0.804 & 0.823 \\
\textit{Pred} & \textit{Known} & 0.898 & 0.896 & 0.897 \\
\textit{Pred} & \textit{Pred} & 0.884 & 0.885 & 0.885 \\
\hline
\end{tabular}
\end{center}
\caption{Reconstruction quality for autoencoders with different decoder architectures: $2^3$ up-convolutions, $4^3$ up-convolutions, and $2^3$ up-convolutions interleaved with $3^3$ convolutions, using different configurations of \textit{Prop-known} and \textit{Prop-pred} propagation modes.}
\label{tbl:technical}
\end{table}

\section{Feature propagation}

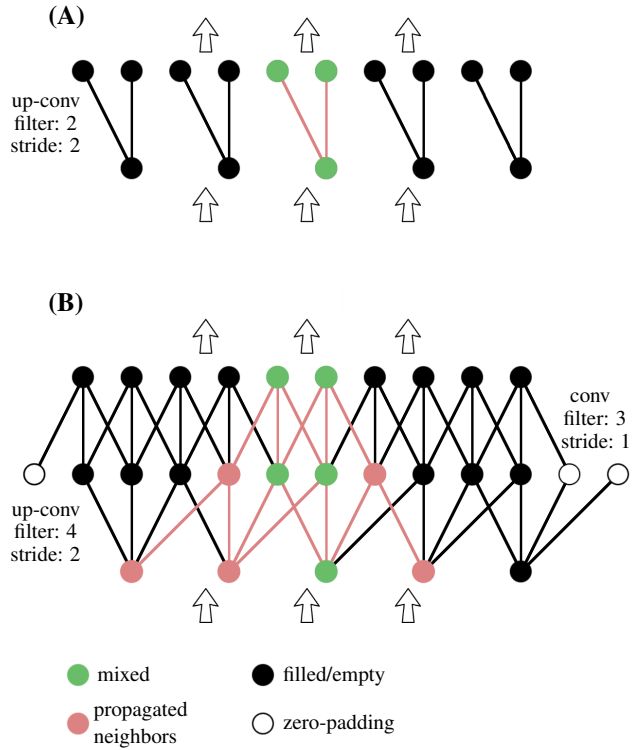
\begin{figure}
\begin{center}
\begin{overpic}[width=\linewidth]{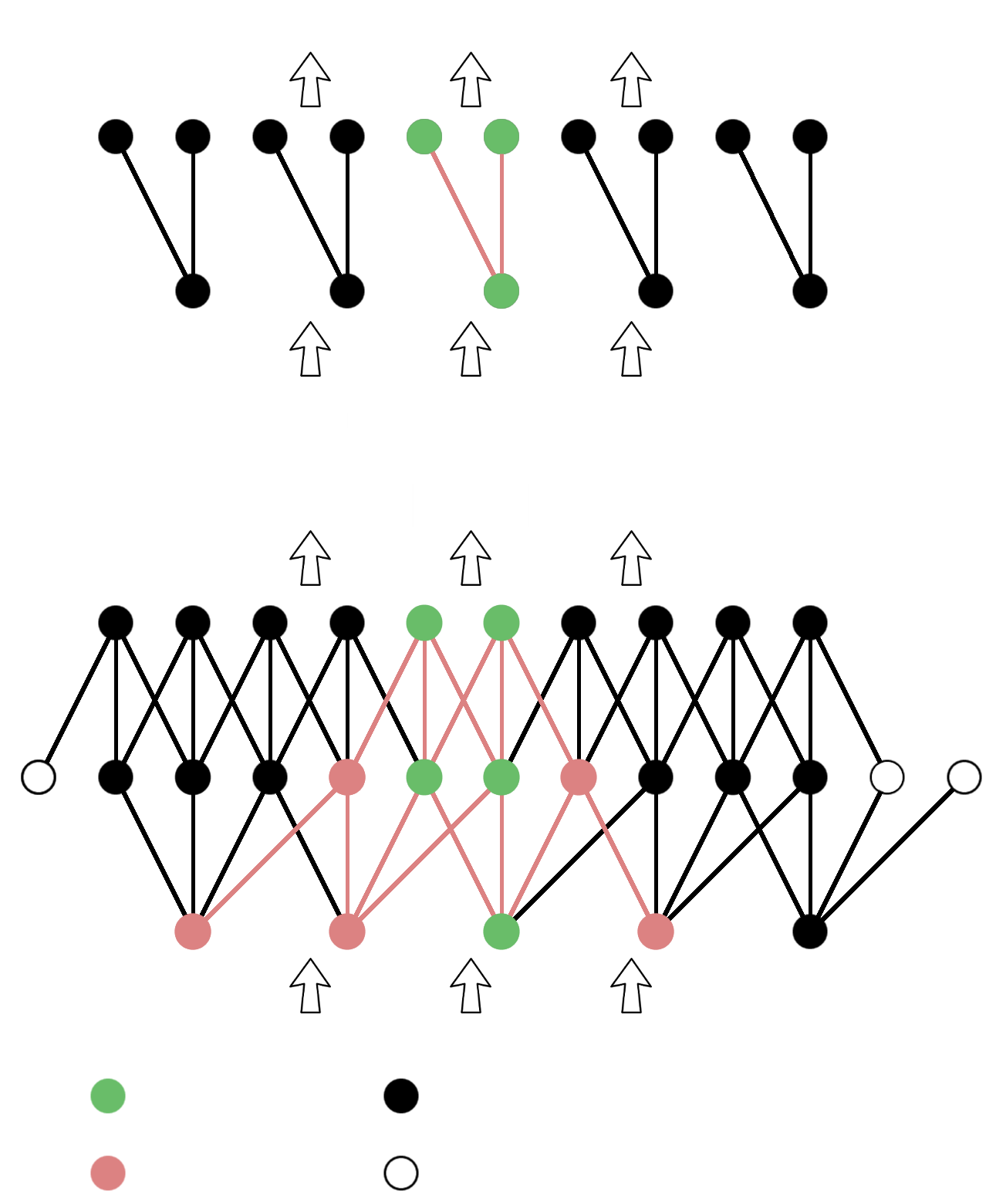}
\put (0.2,84) {\footnotesize up-conv}
\put (0.5,81) {\footnotesize filter: 2}
\put (0,78) {\footnotesize stride: 2}
\put (0.2,30) {\footnotesize up-conv}
\put (0.5,27) {\footnotesize filter: 4}
\put (0,24) {\footnotesize stride: 2}
\put (73.9,45) {\footnotesize conv}
\put (73,42) {\footnotesize filter: 3}
\put (72.7,39) {\footnotesize stride: 1}
\put (36,1.8) {\footnotesize zero-padding}
\put (36,8.2) {\footnotesize filled/empty}
\put (11.5,8.2) {\footnotesize mixed}
\put (11,3.5) {\footnotesize propagated}
\put (11,0) {\footnotesize neighbors}
\put (5,95) {\textbf{(A)}}
\put (5,57) {\textbf{(B)}}
\end{overpic}
\end{center}
   \caption{The \textit{OGNProp} layer propagates the features of "mixed" cells together with the features of the neighboring cells required for computations in subsequent layers. We show the number of neighbors that need to be propagated in two cases: $2^3$ up-convolutions (A), and $4^3$ up-convolutions followed by $3^3$ convolutions (B). Visualized in 1D for simplicity.}
\label{fig:feature_propagation}
\end{figure}

In the main paper we mentioned that the number of features propagated by an \textit{OGNProp} layer depends on the sizes of the convolutional filters in all subsequent blocks.
In case of $2^3$ up-convolutions with stride 2, which were used in most of our experiments, no neighboring features need to be propagated.
This situation is illustrated in Figure~\ref{fig:feature_propagation}-A in a one-dimensional case.
Circles correspond to cells of an octree.
The green cell in the input is the only one for which the value was predicted to be "mixed".
Links between the circles indicate which features of the input are required to compute the result of the operation (convolution or up-convolution) for the corresponding output cell.
In this case, we can see that the output cells in the next level are only affected by their parent cell from the previous level.

A more general situation is shown in Figure~\ref{fig:feature_propagation}-B.
The input is processed with an up-convolutional layer with $4^3$ filters and stride 2, which is followed by a convolutional layer with $3^3$ filters and stride 1.
Again, only one cell was predicted to be "mixed", but in order to perform convolutions and up-convolutions in subsequent layers, we additionally must propagate some of its neighbors (marked red).
Therefore, with this particular filter configuration, two cells in the output are affected by four cells in the input.

Generally, the number of features that should be propagated by each \textit{OGNProp} layer is automatically calculated based on the network architecture before starting the training.

\section{3D shape from high-level information: additional experiments}

\subsection{MPI-FAUST}

\tcn{To additionally showcase the benefit of using higher resolutions, we trained OGNs to fit the MPI-FAUST dataset \cite{faust}.
It contains 300 high-resolution scans of human bodies of 10 different people in 30 different poses.
Same as with the BlendSwap, the trained networks cannot generalize to new samples due to the low amount of training data.}

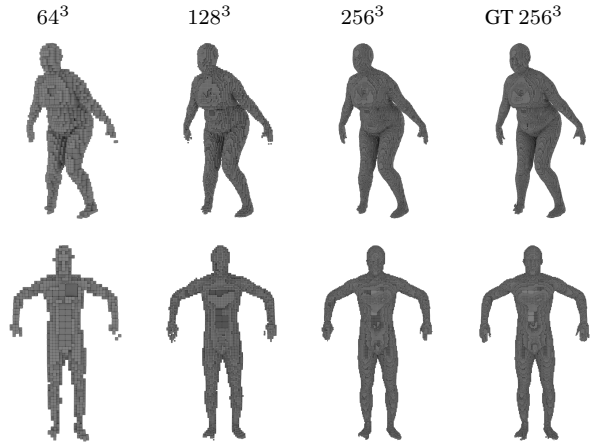
\begin{figure}
	\begin{center}
		\begin{overpic}[width=\linewidth]{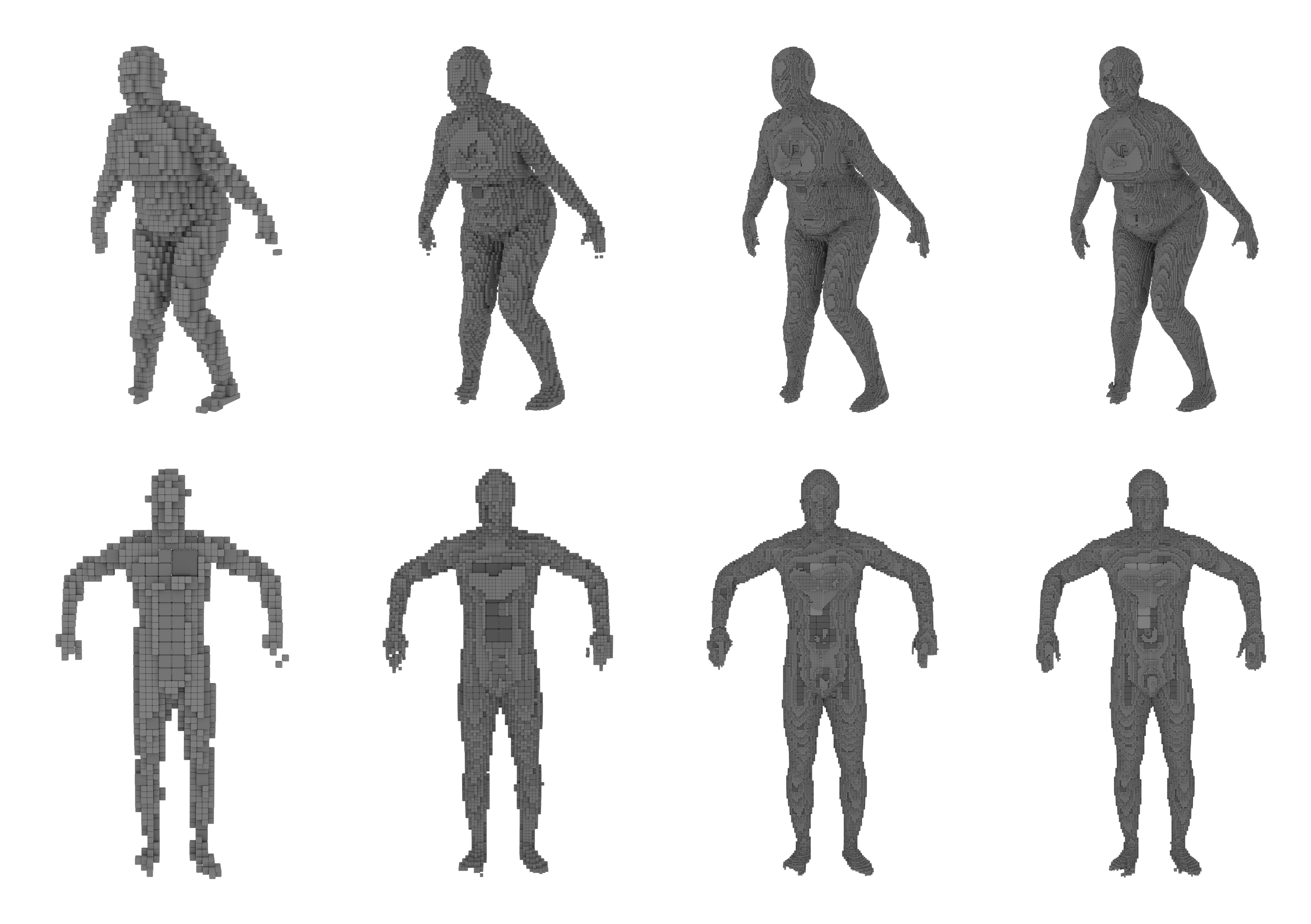}
			\put (8.5,70) {\footnotesize $64^3$}
			\put (32.5,70) {\footnotesize $128^3$}
			\put (57,70) {\footnotesize $256^3$}
			\put (80,70) {\footnotesize GT $256^3$}
		\end{overpic}
	\end{center}
	\caption{Training samples from the FAUST dataset reconstructed by OGN.}
	\label{fig:humans_overfitting}
\end{figure}

\tcn{Figure~\ref{fig:humans_overfitting} and Table~\ref{tbl:faust} demonstrate qualitative and quantitative results respectively.
Human models from MPI-FAUST include finer details than cars from ShapeNet, and therefore benefit from the higher resolution.}

\begin{table}
	\begin{center}
		\begin{tabular}{|c|c|c|}
			\hline
			64 & 128 & 256\\
			\hline
			\hline
			 0.890 & 0.933 & \textbf{0.969}\\
			\hline
		\end{tabular}
	\end{center}
	\caption{3D shape from high-level information on the FAUST dataset. Lower-resolution predictions were upsampled to $256^3$ ground truth.}
	\label{tbl:faust}
\end{table}

\subsection{Fitting reduced ShapeNet-cars}

\begin{figure}
\begin{center}
\begin{overpic}[width=\linewidth]{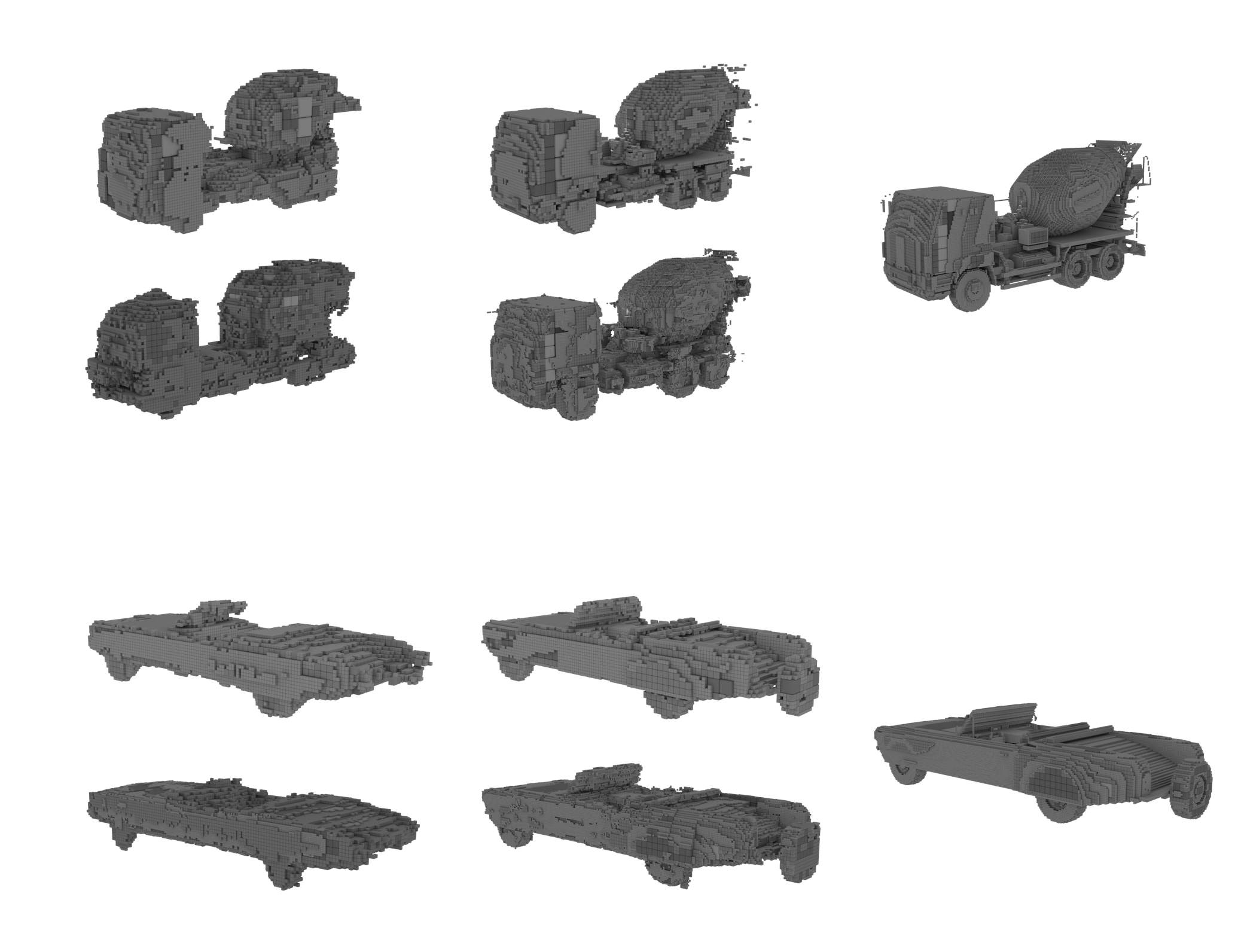}
\put (-1,9) {\footnotesize $256^3$}
\put (-1,23) {\footnotesize $128^3$}
\put (-1,61) {\footnotesize $128^3$}
\put (-1,46) {\footnotesize $256^3$}
\put (17,74) {\footnotesize Full}
\put (47,74) {\footnotesize Subset}
\put (77,74) {\footnotesize GT $256^3$}
\end{overpic}
\end{center}
   \caption{When training on a subset of the Shapenet-cars datset, higher resolution models contain more details.}
\label{fig:shapenet_cars}
\end{figure}
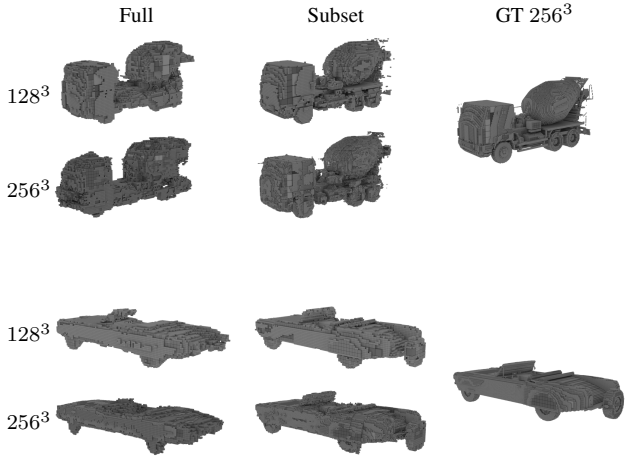

To better understand the performance drop at $256^3$ resolution observed in section 5.4.1 of the main paper, we performed an additional experiment on the ShapeNet-Cars dataset. 
We trained an OGN for generating car shapes from their IDs on a reduced version of ShapeNet-Cars, including just $500$ first models from the dataset.
Quantitative results for different resolutions, along with the results for the full dataset, are shown in Table~\ref{tbl:shapenet_cars}.
Interestingly, when training on the reduced dataset, high resolution is beneficial.
This is further supported by examples shown in Figure~\ref{fig:shapenet_cars}~-- when training on the reduced dataset, the higher-resolution model contain more fine details.
Overall, these results support our hypothesis that the performance drop at higher resolution is not due to the OGN architecture, but due to the difficulty of fitting a large dataset at high resolution.

\begin{table}
\begin{center}
\begin{tabular}{|l|c|c|}
\hline
Dataset & $128^3$ & $256^3$ \\
\hline\hline
Shapenet-cars (full) & 0.901 & 0.865 \\
Shapenet-cars (subset) & 0.922 & 0.931 \\
\hline
\end{tabular}
\end{center}
\caption{There is no drop in performance in higher resolution, when training on a subset of the Shapenet-cars dataset.}
\label{tbl:shapenet_cars}
\end{table}

\section{Shift invariance}

\tcn{The convolution operation on a voxel grid is perfectly shift invariant by design.
This is no longer true for convolutions on octrees: a shift by a single pixel in the original voxel grid can change the structure of the octree significantly.
To study the effect of shifts, we trained two fully convolutional autoencoders - one with an OGN decoder, and one with a dense decoder - on $64^3$ models, with lowest feature map resolution $4^3$ (so the networks should be perfectly invariant to shifts of 16 voxels).
Both were trained on non-shifted Shapenet-Cars, and tested in the \textit{Prop-pred} mode on models shifted by a different number of voxels along the z-axis. The results are summarized in Table~\ref{tbl:shift_invariance}.}

\begin{table}[h]
	\begin{center}
		\begin{tabular}{|c|c|c|}
			\hline
			Shift (voxels) & OGN & Dense\\
			\hline
			\hline
			0 & 0.935 & 0.932\\
			1 & 0.933 & 0.93\\
			2 & 0.929 & 0.925\\
			4 & 0.917 & 0.915\\
			8 & 0.906 & 0.904\\
			\hline
		\end{tabular}
	\end{center} 
	\caption{Fully-convolutional networks tested on shifted data. Even though not shift invariant by design, OGN shows robust performance.}
	\label{tbl:shift_invariance}
\end{table}

\tcn{There is no significant difference between OGN and the dense network.
A likely reason is that different training models have different octree structures, which acts as an implicit regularizer.
The network learns the shape, but remains robust to the exact octree structure.}

\section{Network architectures}

In this section, we provide the exact network architectures used in the experimental evaluations.

\subsection{Autoencoders}

The architectures of OGN autoencoders are summarized in Table~\ref{tbl:autoencoder_architectures}.
For the dense baselines, we used the same layer configurations with usual convolutions instead of \textit{OGNConv}, and predictions being made only after the last layer of the network.
All networks were trained with batch size 16.

\subsection{3D shape from high-level information}

OGN decoders used on the Shapenet-cars dataset are shown in Table~\ref{tbl:architectures}.
Encoders consisted of three fully-connected layers, with output size of the last encoder layer being identical to the input size of the corresponding decoder.

For FAUST and BlendSwap the $256^3$ output octrees had four levels, not five like those in Table~\ref{tbl:architectures}.
Thus, the dense block had an additional deconvolution-convolution layer pair instead of one octree block. The $512^3$ decoder on BlendSwap had one extra octree block with 32 output channels.

All $64^3$ and $128^3$ networks were trained with batch size 16, $256^3$ --- with batch size 4, $512^3$ --- with batch size 1.

\subsection{Single-image 3D reconstruction}

In this experiment we again used decoder architectures shown in Table~\ref{tbl:architectures}.
The architecture of the convolutional encoder is shown in Table~\ref{tbl:encoder_single-view3d}.
The number of channels in the last encoder layer was set identical to the number of input channels of the corresponding decoder.

\begin{table}[h]
\begin{center}
\begin{tabular}{|c|}
\hline
$[137 \times 137 \times 3]$ \\
\hline
Conv ($7 \times 7$) \\
$[69 \times 69 \times 32]$ \\
\hline
Conv ($3 \times 3$) \\
$[35 \times 35 \times 32]$ \\
\hline
Conv ($3 \times 3$) \\
$[18 \times 18 \times 64]$ \\
\hline
Conv ($3 \times 3$) \\
$[9 \times 9 \times 64]$ \\
\hline
Conv ($3 \times 3$) \\
$[5 \times 5 \times 128]$ \\
\hline
FC \\
$[1024]$ \\
\hline
FC \\
$[1024]$ \\
\hline
FC \\
$[4^3 \times c]$ \\
\hline
\end{tabular}
\end{center} 
\caption{Convolutional encoder used in the single-image 3D reconstruction experiment.}
\label{tbl:encoder_single-view3d}
\end{table}

\begin{table*}[h]
\begin{center}
\begin{tabular}{|c||c|c|c|}
\hline
$\mathbf{32^3}$ & $\mathbf{64^3}$ ($2^3$ filters) & $\mathbf{64^3}$ ($4^3$ filters) & $\mathbf{64^3}$ (InvConv) \\
\hline\hline
 & \multicolumn{3}{c|}{$[64^3 \times 1]$} \\
\hline
& \multicolumn{3}{c|}{Conv ($3^3$)} \\
$[32^3 \times 1]$ & \multicolumn{3}{c|}{$[32^3 \times 32]$} \\
\hline
Conv ($3^3$) & \multicolumn{3}{c|}{Conv ($3^3$)} \\
$[16^3 \times 32]$ & \multicolumn{3}{c|}{$[16^3 \times 48]$} \\
\hline
Conv ($3^3$) & \multicolumn{3}{c|}{Conv ($3^3$)} \\
$[8^3 \times 48]$ & \multicolumn{3}{c|}{$[8^3 \times 64]$} \\
\hline
Conv ($3^3$) & \multicolumn{3}{c|}{Conv ($3^3$)} \\
$[4^3 \times 64]$ & \multicolumn{3}{c|}{$[4^3 \times 80]$} \\
\hline
FC & \multicolumn{3}{c|}{FC} \\
$[1024]$ & \multicolumn{3}{c|}{$[1024]$} \\
\hline
FC & \multicolumn{3}{c|}{FC} \\
$[1024]$ & \multicolumn{3}{c|}{$[1024]$} \\
\hline
FC & \multicolumn{3}{c|}{FC} \\
$[4^3 \times 80]$ & \multicolumn{3}{c|}{$[4^3 \times 96]$} \\
\hline
Deconv ($2^3$) & \multicolumn{3}{c|}{Deconv ($2^3$)} \\
$[8^3 \times 64]$ & \multicolumn{3}{c|}{$[8^3 \times 80]$} \\
\hline
Conv ($3^3$)$\rightarrow \mathbf{l1}$ & \multicolumn{3}{c|}{Conv ($3^3$)} \\
$[8^3 \times 64]$ & \multicolumn{3}{c|}{$[8^3 \times 80]$} \\
\hline
\textit{OGNProp} & \multicolumn{3}{c|}{} \\
\textit{OGNConv}$(2^3)\rightarrow \mathbf{l2}$ & \multicolumn{3}{c|}{Deconv ($2^3$)} \\
$[16^3 \times 48]$ & \multicolumn{3}{c|}{$[16^3 \times 64]$} \\
\hline
\textit{OGNProp} & \multicolumn{3}{c|}{} \\
\textit{OGNConv}$(2^3)\rightarrow \mathbf{l3}$ & \multicolumn{3}{c|}{Conv ($3^3$)$\rightarrow \mathbf{l1}$} \\
$[32^3 \times 32]$ & \multicolumn{3}{c|}{$[16^3 \times 64]$} \\
\hline
 & \textit{OGNProp} & \textit{OGNProp} & \textit{OGNProp} \\
 & \textit{OGNConv}$(2^3)\rightarrow \mathbf{l2}$ & \textit{OGNConv}$(4^3)\rightarrow \mathbf{l2}$ & \textit{OGNConv}$(2^3)$ \\
 & $[32^3 \times 48]$ & $[32^3 \times 48]$ & $[32^3 \times 48]$ \\
  &  &  & \textit{OGNConv*}$(3^3)\rightarrow \mathbf{l2}$ \\
 & & & $[32^3 \times 48]$ \\
\hline
 & \textit{OGNProp} & \textit{OGNProp} & \textit{OGNProp} \\
 & \textit{OGNConv}$(2^3)\rightarrow \mathbf{l3}$ & \textit{OGNConv}$(4^3)\rightarrow \mathbf{l3}$ & \textit{OGNConv}$(2^3)$ \\
 & $[64^3 \times 32]$ & $[64^3 \times 32]$ & $[64^3 \times 32]$ \\
  &  &  & \textit{OGNConv*}$(3^3)\rightarrow \mathbf{l3}$ \\
 & & & $[64^3 \times 32]$ \\
\hline
\end{tabular}
\end{center} 
\caption{OGN architectures used in our experiments with autoencoders. \textit{OGNConv} denotes up-convolution, \textit{OGNConv*} --- convolution. Layer name followed by '$\rightarrow \mathbf{lk}$' indicates that level $k$ of an octree is predicted by a classifier attached to this layer.}
\label{tbl:autoencoder_architectures}
\end{table*}

\begin{table*}[h]
\begin{center}
\begin{tabular}{|c|c|c|c|}
\hline
$\mathbf{32^3}$ & $\mathbf{64^3}$ & $\mathbf{128^3}$ & $\mathbf{256^3}$ \\
\hline\hline
$[4^3 \times 80]$ & $[4^3 \times 96]$ & $[4^3 \times 112]$ & $[4^3 \times 112]$ \\
\hline
Deconv ($2^3$) & Deconv ($2^3$) & Deconv ($2^3$) & Deconv ($2^3$) \\
$[8^3 \times 64]$ & $[8^3 \times 80]$ & $[8^3 \times 96]$ & $[8^3 \times 96]$ \\
\hline
Conv ($3^3$) $\rightarrow \mathbf{l1}$ & Conv ($3^3$) & Conv ($3^3$) & Conv ($3^3$) \\
$[8^3 \times 64]$ & $[8^3 \times 80]$ & $[8^3 \times 96]$ & $[8^3 \times 96]$ \\
\hline
\textit{OGNProp} & & & \\
\textit{OGNConv} ($2^3$) $\rightarrow \mathbf{l2}$ & Deconv ($2^3$) & Deconv ($2^3$) & Deconv ($2^3$)\\
$[16^3 \times 48]$ & $[16^3 \times 64]$ & $[16^3 \times 80]$ & $[16^3 \times 80]$ \\
\hline
\textit{OGNProp} & & & \\
\textit{OGNConv} ($2^3$) $\rightarrow \mathbf{l3}$ & Conv ($3^3$) $\rightarrow \mathbf{l1}$ & Conv ($3^3$) $\rightarrow \mathbf{l1}$ & Conv ($3^3$) $\rightarrow \mathbf{l1}$ \\
$[32^3 \times 32]$ & $[16^3 \times 64]$ & $[16^3 \times 80]$ & $[16^3 \times 80]$ \\
\hline
& \textit{OGNProp} & \textit{OGNProp} & \textit{OGNProp}\\
& \textit{OGNConv} ($2^3$) $\rightarrow \mathbf{l2}$ & \textit{OGNConv} ($2^3$) $\rightarrow \mathbf{l2}$ & \textit{OGNConv} ($2^3$) $\rightarrow \mathbf{l2}$\\
& $[32^3 \times 48]$ & $[32^3 \times 64]$ & $[32^3 \times 64]$ \\
\hline
& \textit{OGNProp} & \textit{OGNProp} & \textit{OGNProp}\\
& \textit{OGNConv} ($2^3$) $\rightarrow \mathbf{l3}$ & \textit{OGNConv} ($2^3$) $\rightarrow \mathbf{l3}$ & \textit{OGNConv} ($2^3$) $\rightarrow \mathbf{l3}$ \\
& $[64^3 \times 32]$ & $[64^3 \times 48]$ & $[64^3 \times 48]$ \\
\hline
&  & \textit{OGNProp} & \textit{OGNProp} \\
&  & \textit{OGNConv} ($2^3$) $\rightarrow \mathbf{l4}$ & \textit{OGNConv} ($2^3$) $\rightarrow \mathbf{l4}$ \\
&  & $[128^3 \times 32]$ & $[128^3 \times 32]$ \\
\hline
& & & \textit{OGNProp}\\
&  & & \textit{OGNConv} ($2^3$) $\rightarrow \mathbf{l5}$ \\
&  & & $[256^3 \times 32]$ \\
\hline
\end{tabular}
\end{center} 
\caption{OGN decoder architectures used in shape from ID, and single-image 3D reconstruction experiments.}
\label{tbl:architectures}
\end{table*}

\end{document}